\documentclass[letterpaper, 10 pt, conference]{ieeeconf}  %

\IEEEoverridecommandlockouts                              %

\usepackage[table]{xcolor}
\usepackage{comment}
\usepackage{pifont}
\usepackage{graphics} %
\usepackage{booktabs} %
\usepackage{multirow}
\usepackage{epsfig} %
\usepackage{mathptmx} %
\usepackage{times} %
\usepackage{amsmath} %
\usepackage{amssymb}  %
\usepackage{graphicx}
\usepackage{wrapfig}
\usepackage[labelfont=bf]{caption}
\usepackage{subcaption}
\usepackage{lipsum} 
\usepackage{url} 
\usepackage{hyperref}
\usepackage{listings}

\newcommand\blfootnote[1]{%
  \begingroup
  \renewcommand\thefootnote{}\footnote{#1}%
  \addtocounter{footnote}{-1}%
  \endgroup
}

\definecolor{codegreen}{rgb}{0,0.6,0}
\definecolor{codegray}{rgb}{0.5,0.5,0.5}
\definecolor{codepurple}{rgb}{0.58,0,0.82}
\definecolor{backcolour}{rgb}{0.95,0.95,0.92}
\lstdefinestyle{style-prompt}{
    basicstyle=\ttfamily\footnotesize,
    backgroundcolor=\color{backcolour},
    basewidth=0.55em,
    commentstyle=\color{codegreen},
    breakatwhitespace=false,
    breaklines=true,
    captionpos=b,
    keepspaces=true,
    showspaces=false,
    showstringspaces=false,
    keywordstyle=\ttfamily,
    numberstyle=\tiny\color{codegray},
    numbers=left,
    numbersep=5pt, 
    moredelim=[is][\color{red}]{|}{|},
    moredelim=[is][\color{nice-blue}\bfseries]{//}{//},
    moredelim=[is][\color{nice-orange}\bfseries]{[[}{]]}
}

\title{\LARGE \bf
Towards Generalizable Vision-Language Robotic Manipulation:\\ A Benchmark and LLM-guided 3D Policy
}

\author{Ricardo Garcia$^{*\dagger}$, Shizhe Chen$^{*\dagger}$, Cordelia Schmid$^\dagger$%
\thanks{$^*$The authors contributed equally to this work.}%
\thanks{$^{\dagger}$Inria, École normale supérieure, CNRS, PSL Research University
        {\tt\small {firstname.lastname}@inria.fr}}%
}

\begin{document}

\twocolumn[{%
  \renewcommand\twocolumn[1][]{#1}%
  \maketitle

  \centering
  \includegraphics[width=1\linewidth]{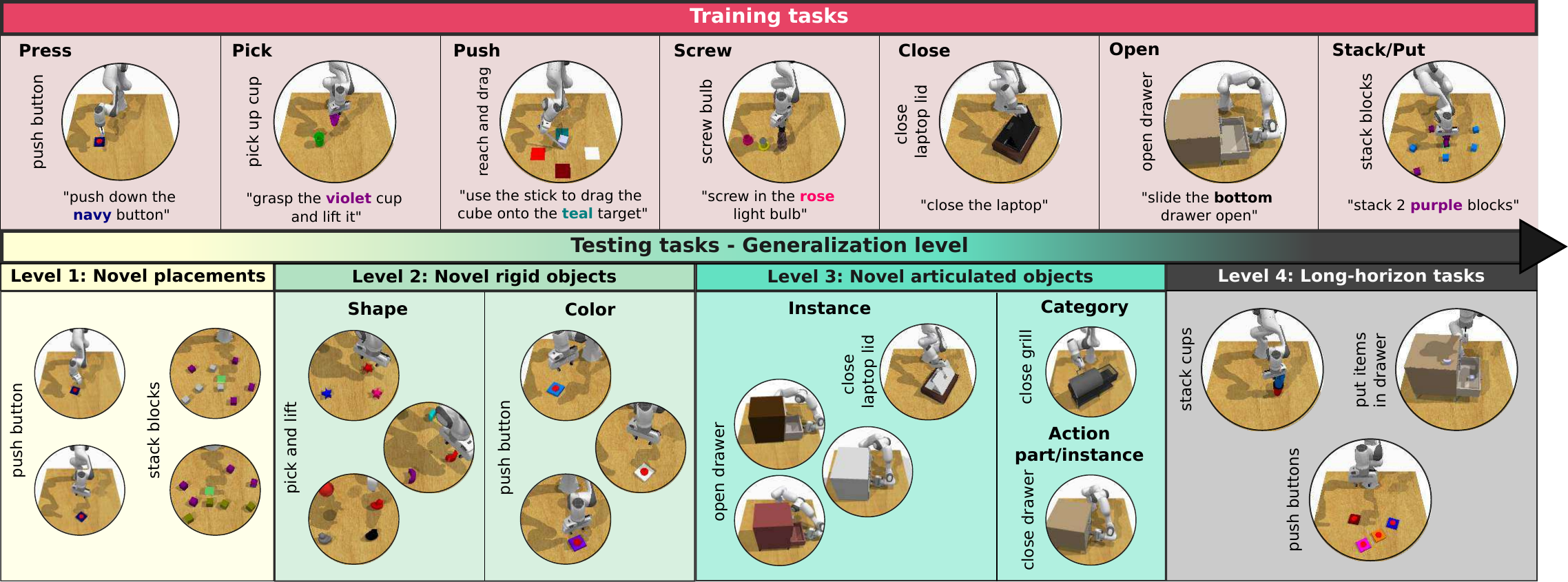}
  \begin{center}
  \captionof{figure}{\textbf{GemBench benchmark for vision-language robotic manipulation.} Top: GemBench comprises 16 training tasks with 31 variations, covering seven action primitives. Bottom: The testing set includes 44 tasks with 92 variations, which are organized into four progressively more challenging levels to systematically evaluate generalization capabilities.}
  \label{fig:benchmark}
  \end{center}
}]

\thispagestyle{empty}
\pagestyle{empty}

\begin{abstract}

Generalizing language-conditioned robotic policies to new tasks remains a significant challenge, 
hampered by the lack of suitable simulation benchmarks.
In this paper, we address this gap by introducing GemBench, a novel benchmark to assess  generalization capabilities of vision-language robotic manipulation policies. 
GemBench incorporates seven general action primitives and four levels of generalization, spanning novel placements, rigid and articulated objects, and complex long-horizon tasks.
We evaluate state-of-the-art approaches on GemBench and also introduce a new method. 
Our approach 3D-LOTUS leverages rich 3D information for action prediction conditioned on language. While 3D-LOTUS excels in both efficiency and performance on seen tasks, it struggles with novel tasks.
To address this,  we present 3D-LOTUS++, a framework that integrates 3D-LOTUS's motion planning capabilities with the task planning capabilities of LLMs and the object grounding accuracy of VLMs.
3D-LOTUS++ achieves state-of-the-art performance on novel tasks of GemBench, setting a new standard for generalization in robotic manipulation. 
The benchmark, codes and trained models are available at \url{https://www.di.ens.fr/willow/research/gembench/}.%
\blfootnote{$^*$The authors contributed equally to this work.}%
\blfootnote{$^{\dagger}$Inria, École normale supérieure, CNRS, PSL Research University
        {\tt\small {firstname.lastname}@inria.fr}}%
        
\end{abstract}

\section{Introduction}

Vision-language robotic manipulation aims to train policies performing complex tasks based on visual inputs and language instructions~\cite{brohan2022rt1,chen2023polarnet,huang2023voxposer,gervet2023act3d}.
Given the vast diversity of real-world tasks, collecting data for every possible task is prohibitively expensive. Hence, it is critical to develop models that can effectively generalize to novel tasks.%

A number of approaches have been proposed to improve the generalization ability of robotic policies. 
One prominent direction focuses on pretraining visual and language representations on large-scale web data for robotics~\cite{nair2022r3m,radosavovic2023mvp,karamcheti2023voltron,chen2024sugar}. While these representations are more powerful, they do not directly translate into generalizable policies for new tasks.
Another approach~\cite{brohan2022rt1,brohan2023rt2,driess2023palme,team2024octo,kim2024openvla} pretrains entire robotic policies by combining robot data with web data, but these methods remain constrained by the limited size and diversity of available robot datasets~\cite{vuong2023openx}.
More recently, foundation models such as large language models (LLMs)~\cite{llama3modelcard} and vision-language models (VLMs)~\cite{kirillov2023sam} have been employed to enhance generalization of robot policies. For example, CaP~\cite{liang2023codepolicy} uses LLMs to generate codes that executes primitive actions, and VoxPoser~\cite{huang2023voxposer} composes value maps with foundation models for action execution with classic motion planning. However, these approaches focus on simpler tasks like pick-and-place and often struggle with more complex manipulation like rotating objects.

\begin{table*}
\centering
\small
\tabcolsep=0.1cm
\caption{\textbf{Comparison of benchmarks for vision-and-language robotic manipulation.} \textit{Multi-skill:} covering multiple action primitives beyond pick and place. \textit{atc-obj:} tasks involve interactions with articulated objects. \textit{Test generalization level to attr-obj:} unseen size, color or texture of the object, \textit{act-obj:} unseen action object combination, \textit{inst} or \textit{cate}: same action for unseen object instance or category respectively and \textit{long-horizon:} unseen combination of multiple seen actions and objects.}
\label{tab:benchmark_cmpr}
\begin{tabular}{llccccccccccc} \toprule
\multirow{2}{*}{Benchmark} & \multirow{2}{*}{Simulator} & \multirow{2}{*}{Physics} & \multirow{2}{*}{\begin{tabular}[c]{@{}c@{}}\# Train \\ task(var)\end{tabular}} & \multirow{2}{*}{\begin{tabular}[c]{@{}c@{}}\# Test \\ task(var)\end{tabular}} & \multirow{2}{*}{\begin{tabular}[c]{@{}c@{}}Multi- \\ skills\end{tabular}} & \multirow{2}{*}{\begin{tabular}[c]{@{}c@{}}atc- \\ obj\end{tabular}} & \multicolumn{5}{c}{Test generalization level} \\
 &  &  &   &  &  & & attr-obj & act-obj & inst & cate & long-horizon \\ \midrule
 RLBench-74Task~\cite{guhur2023hiveformer} & RLBench & \ding{51} & 74 (74) & 74 (74) & \ding{51} & \ding{51} &  \ding{55} & \ding{55} & \ding{55} & \ding{55} & \ding{55} \\
RLBench-18Task~\cite{shridhar2023peract} & RLBench & \ding{51} & 18 (249)  & 18 (249) & \ding{51} & \ding{51} & \ding{55} & \ding{55} & \ding{55} & \ding{55} & \ding{55} \\
VLMBench~\cite{zheng2022vlmbench} & RLBench & \ding{51} & 8 (233) & 8 (374) & \ding{51} & \ding{51} & \ding{51} &  \ding{55} &  \ding{51} & \ding{55} & \ding{55} \\
ALFRED~\cite{shridhar2020alfred} & AI2-THOR & \ding{55} & 7 (21,023) & 7 (1,529) & \ding{51} & \ding{51} &  \ding{51} & \ding{55} & \ding{55} & \ding{55} & \ding{51} \\
Calvin~\cite{mees2022calvin} & PyBullet  & \ding{51} & 34 & 34 (1000) & \ding{51} & \ding{51} & \ding{51} & \ding{55} & \ding{55} & \ding{55} & \ding{51} \\
Ravens~\cite{zeng2020transporter} & PyBullet & \ding{51}  & 10  & 10  & \ding{55} & \ding{55} & \ding{51} & \ding{55} & \ding{51} & \ding{51}  & \ding{55} \\
Arnold~\cite{gong2023arnold} & Isaac Sim & \ding{51} & 8 (3571) & 8 (800) & \ding{51} & \ding{51} & \ding{51} & \ding{51} & \ding{51} & \ding{55}  & \ding{55} \\
VIMA-Bench~\cite{jiang2023vima} & Ravens & \ding{51} & 13 & 17 & \ding{55} & \ding{55} & \ding{51} & \ding{55} & \ding{51} & \ding{51} & \ding{51} \\
Colosseum~\cite{pumacay2024colosseum} & RLBench & \ding{51} & 20 (280) & 20 (20,371) & \ding{51} & \ding{51} & \ding{51} & \ding{55} & \ding{55} & \ding{55} & \ding{55} \\
GemBench (Ours) & RLBench & \ding{51} & 16 (31) & 44 (92) & \ding{51} & \ding{51} &  \ding{51} & \ding{51} & \ding{51} & \ding{51} & \ding{51} \\ \bottomrule
\end{tabular}
\end{table*}

Furthermore, there is a notable lack of systematic benchmarks to evaluate the generalization capabilities of models. 
Though several simulation-based benchmarks have been developed for vision-language robotic manipulation~\cite{james2020rlbench,mees2022calvin,zheng2022vlmbench}, they primarily evaluate models on tasks seen during training, neglecting the crucial aspect of generalization.
A few exceptions like VIMA-Bench~\cite{jiang2023vima} and Colosseum~\cite{pumacay2024colosseum} attempt to evaluate generalization abilities. However, VIMA-Bench is limited to relatively simple action skills, while Colosseum focuses more on environment perturbations like lighting changes rather than generalizing to new tasks.

In this work, we introduce {\bf GEM}Bench - a new simulation benchmark to evaluate {\bf GE}neralization capabilities for vision-language robotic {\bf M}anipulation.
As shown in Figure~\ref{fig:benchmark}, GemBench features two key improvements compared to prior work.
First, it incorporates a wider range of complex tasks, involving seven core action skills: \emph{press, pick, push, screw, close, open} and \emph{put}.
Secondly, it introduces a comprehensive suite of four generalization levels with increasing difficulty, focusing on generalization to novel placements, rigid objects, articulated objects, and long-horizon tasks respectively.

To tackle the problem, we first propose a new {\bf 3D} robotic manipulation policy - {\bf 3D-LOTUS} with {\bf L}anguage-c{\bf O}nditioned poin{\bf T} clo{\bf U}d tran{\bf S}former. 
By leveraging a strong 3D backbone and an effective action representation, 3D-LOTUS achieves state-of-the-art performance on existing vision-language manipulation benchmark~\cite{shridhar2023peract} and Level 1 of GemBench, while significantly improving training efficiency.
Nevertheless, 3D-LOTUS struggles to generalize to new tasks in GemBench, mainly due to limitations in planning for new tasks and grounding new objects.
Therefore, we propose an enhanced version {\bf 3D-LOTUS++} which integrates foundation models to boost generalization capabilities. 
Specifically, LLMs are employed for task planning, decomposing tasks into step-by-step actionable plans, and VLMs are used for object grounding which can localize new objects mentioned in the plan. With the grounded object and the primitive action in the plan, 3D-LOTUS serves as the motion controller to generate action trajectories.
Experimental results demonstrate that 3D-LOTUS++ significantly improves generalization, outperforming 3D-LOTUS on Levels 2 to 4 of GemBench.

\noindent To summarize, our contributions are three fold:

\noindent \textbf{$\bullet$}  We introduce a new benchmark GemBench to systematically evaluate generalization in vision-language robotic manipulation across four complexity levels.

\noindent \textbf{$\bullet$} We propose an effective manipulation policy 3D-LOTUS and enhance its generalization ability with foundation models for task planning and object grounding (3D-LOTUS++). 

\noindent \textbf{$\bullet$} Our models establish state of the arts on existing benchmark and GemBench, and also work reliably on a real robot.

\section{Related work}
\label{sec:related}

\noindent \textbf{Robotic manipulation benchmark.}
Significant progress has been made in the development of robot simulators such as RLBench~\cite{james2020rlbench}, AI2-THOR~\cite{kolve2017ai2thor} and Isaac Sim~\cite{makoviychuk2021isaacgym}.
Leveraging these simulators, various benchmarks have emerged for robotic manipulation.
Early benchmarks~\cite{guhur2023hiveformer,shridhar2023peract} train and test policies on the same task set, overlooking the critical aspect of generalization to unseen scenarios. 
To address this, more recent benchmarks~\cite{zheng2022vlmbench,shridhar2020alfred,mees2022calvin,zeng2020transporter,gong2023arnold,pumacay2024colosseum,jiang2023vima} have introduced generalization evaluations on new compositions of objects and colors, new object shapes, or even long-horizon tasks.
Among them, VIMA-Bench~\cite{jiang2023vima} and Colosseum~\cite{pumacay2024colosseum} are most similar to our work, aiming to systematically evaluate different generalization abilities.
However, VIMA-Bench~\cite{jiang2023vima} is limited to pick-and-place tasks using a suction gripper, while Colosseum~\cite{pumacay2024colosseum} emphasizes generalization to environment perturbations such as changes in lighting and camera angles. 
In contrast, GemBench covers more complex action skills and evaluate generalization to entirely new tasks rather than perturbations of the seen tasks.
Table~\ref{tab:benchmark_cmpr} provides a comprehensive comparison of these benchmarks. 
We can observe that GemBench is most general among all of them.

\noindent \textbf{Vision-and-language robotic manipulation.}
Learning robotic manipulation conditioned on vision and language has received increasing attention~\cite{shao2021concept2robot,lynch2023interactivelang,stepputtis2020language}. 
The high dimensionality of manipulation action space makes it challenging to directly use reinforcement learning (RL) in training~\cite{kalashnikov2018scalablerl}. Therefore, most works rely on imitation learning~(IL)~\cite{jang2022bcz,brohan2022rt1,guhur2023hiveformer,shridhar2023peract,goyal2023rvt,chen2023polarnet,chen2024sugar,gervet2023act3d,ke20243ddifusseractor} using scripted trajectories~\cite{james2020rlbench} or tele-operation data~\cite{vuong2023openx}. 
Visual representation plays a crucial role in policy learning. Existing works~\cite{jang2022bcz,brohan2022rt1,chi2023diffusionpolicy,guhur2023hiveformer,goyal2023rvt,goyal2024rvt2} mainly use 2D images to predict actions, though recent works have begun exploring 3D representations~\cite{james2022c2farm,shridhar2023peract,chen2023polarnet,gervet2023act3d,ke20243ddifusseractor,chen2024sugar}.
In this work, we take advantage of rich spatial information of 3D point cloud for motion planning, while leaving object grounding in 2D to benefit from the generalization strength of pretrained 2D models~\cite{kirillov2023sam,radford2021clip,minderer2023owlv2}.

\noindent \textbf{Foundation models for robotics.}
Learning-based robotic policies often struggle to generalize to new scenarios~\cite{yu2023scaling}.
Inspired by the remarkable generalization capabilities of foundation models~\cite{radford2021clip,kirillov2023sam,openai2023gpt4}, recent work  explores how to leverage these models for planning, perception and control in robotics.
Huang \emph{et al.}~\cite{huang2022languageplanner} use LLMs to decompose high-level tasks into sub-steps. To ground plans in the visual world, SayCan~\cite{brohan2023saycan} combine LLMs with value functions of pretrained skills. ViLa~\cite{hu2023vila} replaces LLMs with a multimodal LLM GPT-4V~\cite{openai2023gpt4v}. 
CaP~\cite{liang2023codepolicy} instructs LLMs to write code which call tools for perception and control.
However, these approaches rely on predefined motion skills, limiting  applicability to broader tasks. 
To address this, VoxPoser~\cite{huang2023voxposer} use LLMs to construct 3D voxel maps of affordance, constraint, rotation and velocity, which are fed into traditional motion planing algorithms to plan a trajectory.
Nevertheless, VoxPoser only provides a coarse-grained understanding of the scene and struggles with precise robot control.
In this work, we propose to combine the generalization ability of foundation models with strong motion control capabilities of 3D policies for robotic manipulation.

\section{\textbf{GEM}Bench: \textbf{GE}neralizable Vision-Language Robotic \textbf{M}anipulation Benchmark}
\label{sec:benchmark}

This paper introduces the GemBench benchmark to systematically evaluate generalization capabilities of vision-and-language robotic manipulation policies. It is built upon the RLBench simulator~\cite{james2020rlbench}, which provides a wide range of visually and physically realistic tasks together with a framework to generate scripted demonstrations.
In the following, we first describe training tasks in GemBench in Sec~\ref{sec:benchmark_train} and then present four levels of generalization for evaluation in Sec~\ref{sec:benchmark_test}.
The details of the proposed GemBench are presented in Sec~\ref{sec:suppmat_gembench} in the appendix.

\subsection{Training tasks}
\label{sec:benchmark_train}

We select 16 tasks (31 variations\footnote{The RLBench tasks typically involve manipulating fixed-shape objects, such as pressing a button, with variations including different object colors (e.g., red or blue button), parts (e.g., top or middle drawer), and manipulation sequences (e.g., press a red button followed by a green one).}) from existing RLBench benchmarks~\cite{james2020rlbench, shridhar2023peract} to capture a diverse range of action primitives beyond simple pick-and-place. These tasks, shown in Figure~\ref{fig:benchmark} (top), include seven action primitives: press, pick, push, screw, close, open, and stack/put.
Examples of training tasks are \emph{push button, pick up cup, reach and drag cube, screw light bulb in,  close laptop lid, open drawer, stack blocks.}
Task variations cover 20 objects (e.g., cube, cup, fridge), 20 colors (e.g., red, blue, violet), and 3 object parts (e.g., top, middle, bottom). 
The training set is sufficiently diverse and should enable a robot to generalize to new tasks, such as novel attribute-object compositions, new action-object pairings, or even entirely new shapes.

\subsection{Testing tasks with four levels of generalization}
\label{sec:benchmark_test}

As shown in Figure~\ref{fig:benchmark} (bottom), the test set includes four levels of 
generalization that progressively increase the difficulty of vision-language robotic manipulation.
The test set consists of a total of 44 tasks (92 variations), with 23 tasks selected from the original RLBench 100 tasks \cite{james2020rlbench} and 21 newly scripted tasks. 
The test levels differ in object shape and color, object articulation and  horizon of the tasks.

\noindent \textbf{Level 1 - Novel placements:} 
This level consists of the same 16 tasks (31 variations) as the training set, but with new object placements randomly sampled within the robot's workspace. Additionally, some tasks feature new distractor objects with different colors. 
The objective is to evaluate whether a policy can perform well on seen tasks with minor configuration changes.\\
\noindent \textbf{Level 2 - Novel rigid objects:} 
This level comprises 15 unseen tasks (28 variations) where the robot interacts with novel rigid objects using actions such as press, pick, and put. %
There are two categories for generalization to rigid objects:
1) Novel object-color compositions. For instance, training tasks only manipulate yellow button and rose bulb, while the test task requires to operate a rose button. This level includes 20 new object-color compositions.
2)~Novel object shapes. For example, picking a cube is learned in training, but the testing task is  lifting a toy or a star-shaped item. There are 8 new object shapes in the evaluation.
 
\noindent \textbf{Level 3 - Novel articulated objects}: 
    This set includes 18 new tasks (21 variations) where the robot interacts with articulated objects.
    Three categories are proposed:
1) Novel action-part compositions. For example, if trained to open bottom drawer and put item in middle shelf, it now needs to open middle drawer.
There are  8 novel compositions.
2) Novel instances. 
For example, after being trained on a three-drawer unit, it must generalize to one with four drawers. This set includes 11 new object instances.
3) Novel categories. For example, it is trained to close a laptop lid but must generalize to closing a grill lid. This includes 2 new object categories.

\noindent \textbf{Level 4 - Novel Long-horizon tasks}: This level presents the greatest challenge as it requires the robot to combine multiple actions learned during training. It includes 6 long-horizon tasks (12 variations). For example, the task ``put items in drawer" involves a sequence of actions such as opening the drawer, picking up a sequence of items (a cube, a cylinder and a moon), and placing them inside according to the order specified by the variation instruction.

\section{Method}
\label{sec:method}

\begin{figure*}[tp]
    \centering    \includegraphics[width=0.95\linewidth]{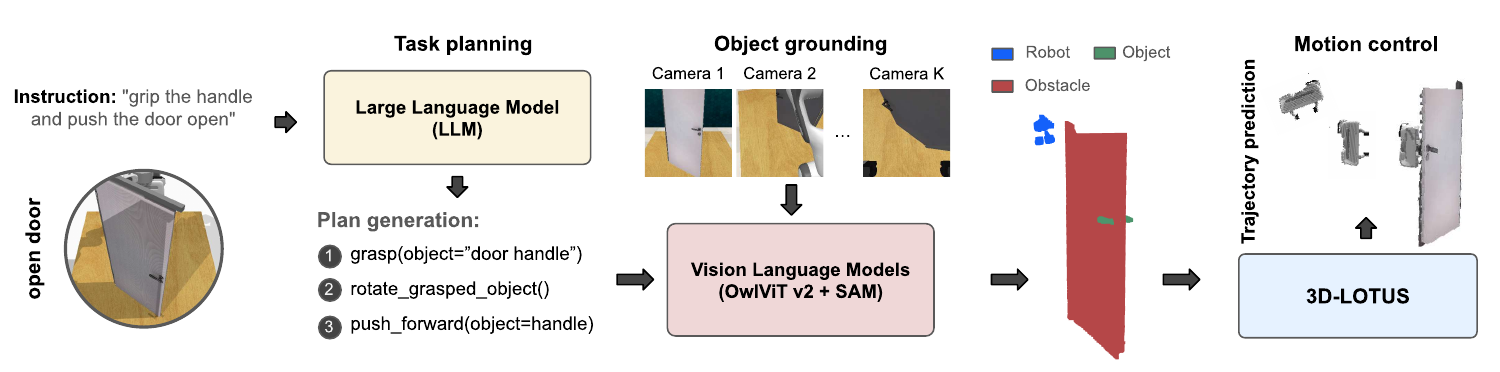}
    \caption{\textbf{Overview of 3D-LOTUS++ framework.} It leverages generalization capabilities of foundation models for planning and perception, and strong action execution ability of 3D-LOTUS to perform complex tasks.}
    \label{fig:method_overview}
\end{figure*}

\subsection{Problem formulation}
\label{sec:method_problem}

The goal is to learn a policy $\pi (a_t | O_t, L)$ for robotic manipulation, where $L$ is a language instruction, $O_t \in \mathbb{O}, a_t \in \mathbb{A}$ are the observation and action at step $t$ respectively, with $\mathbb{O}$ and $\mathbb{A}$ denoting the observation and action space.

The observation space $\mathbb{O}$ includes aligned RGB-D images from $K$ cameras, along with the robot's proprioceptive state consisting of joint and gripper poses.
We assume the intrinsic and extrinsic camera parameters are known. 

The action space $\mathbb{A}$ comprises the gripper's position $a^{p}_t \in \mathbb{R}^{3}$, rotation $a^{r}_t \in \mathbb{R}^{3}$, and open state $a^{o}_{t} \in \{0, 1\}$ indicating if the gripper is open or closed.
We utilize waypoint representation~\cite{james2022c2farm,liu22autolambda,guhur2023hiveformer,shridhar2023peract} for action sequences. Inverse kinematics is used to move the robot from $a_{t-1}$ to $a_t$.

\subsection{3D-LOTUS policy}
\label{sec:method_policy}

\begin{figure}[tp]
    \centering    \includegraphics[width=0.95\linewidth]{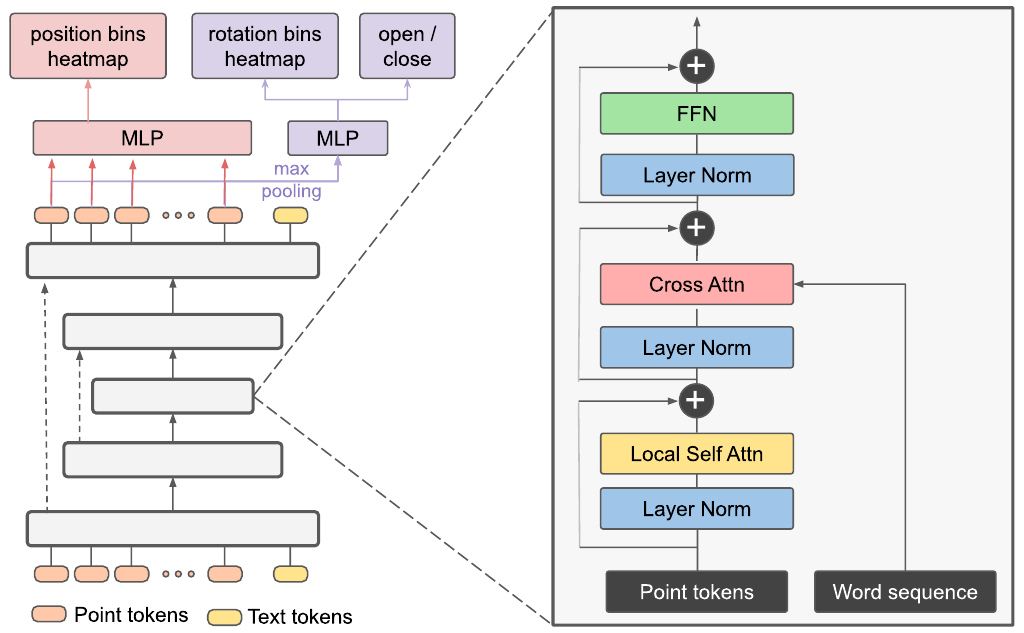}
    \caption{\textbf{3D-LOTUS architecture.} It takes point cloud and text as input to predict the next action.}
    \label{fig:3dlotus_overview}
\end{figure}

As 3D point clouds provide rich spatial information to perceive object shapes and positions, we propose to effectively exploit the 3D information for robotic manipulation.
The 3D-LOTUS policy is a \textbf{3D} robotic manipulation policy with \textbf{l}anguage-c\textbf{o}nditioned poin\textbf{t} clo\textbf{u}d tran\textbf{s}former. 
Figure~\ref{fig:3dlotus_overview} illustrates the architecture of the policy.

\noindent \textbf{Point cloud preprocessing.}
We follow PolarNet~\cite{chen2023polarnet} to project multi-view RGB-D images into a unified point cloud in world coordinates, then downsample this point cloud to one point per 1cm$^3$ voxel~\cite{Zhou2018open3d}.
We exclude points outside the robot's workspace, and points on the robotic arm using a CAD model and joint poses of the arm, as these contribute little for manipulation.
The resulting point cloud $V$ of $n$ points only covers the objects and robot gripper, significantly reducing the number of points and improving speed without compromising performance.
Each point $v_i \in V$ consists of XYZ coordinates $v^p_i$ and additional  feature $v^o_i$ such as RGB color and relative height to the table.

\noindent \textbf{Language-conditioned point cloud transformer.} 
We employ point cloud transformer v3 (PTV3)~\cite{wu2024ptv3} as backbone to encode point cloud $V$. PTV3 adopts a U-Net~\cite{ronneberger2015unet} architecture with downsampling and upsampling blocks to efficiently compute point embeddings, where each block consists of transformer layers. For more details, please refer to the PTV3 paper~\cite{wu2024ptv3}.
We explore two variants to incorporate language information into the PTV3 model.
Assuming the language instruction $L$ is encoded by a CLIP text encoder~\cite{radford2021clip} and represented as a sequence of word embeddings $(w_1, \cdots, w_L)$.
The first variant employs the \emph{adaptive normalization approach}~\cite{dubey2023adanorm}. We compute a global language embedding $\overline{w}$ by weighted averaging $(w_1, \cdots, w_L)$, and use $\overline{w}$ to directly regress dimension-wise scale and shift parameters for each normalization layer in PTV3. 
The second variant utilizes a more conventional \emph{cross-attention mechanism}~\cite{vaswani2017attention}. A cross-attention layer is added after each self-attention layer in PTV3, enabling each point to attend to the entire sequence of word embeddings $(w_1, \cdots, w_L)$. 

\noindent \textbf{Action prediction.}
Let $v^e_i$ denote the final point embedding for each point $v_i$ after the language-conditioned PTV3 model.
We propose a new classification-based approach for action prediction, in contrast to the regression-based approach~\cite{guhur2023hiveformer,chen2023polarnet,chen2024sugar} or inefficient position classification over the whole 3D workspace~\cite{shridhar2023peract,gervet2023act3d}.
For position prediction, we predict the gripper's location along the $X, Y, Z$ axes separately.
We define sequential bins $v_{i,k,j}$ centered at each point's location $v^p_i$ for each axis $k \in \{X, Y, Z\}$, with $j \in [-m, m]$ representing the bin index. Each bin has a size of $b$, so the position of bin $v_{i,k,j}$ along $k$-axis is given by $v^p_{i,k} + b \times j$.
Using $v^e_i$, we predict a heatmap for these bins at each point, and concatenate the bins across all points to form the final heatmap for each axis.
During inference, we select the bin with the highest probability to determine the position for each axis.
For rotation prediction, we also discretize the Euler angle for each axis into bins and use classification to predict the angles. 
The open state prediction remains a binary classification task.
We apply max pooling over all points $v^e_i$ to predict rotation and open state.
The cross entropy loss is employed to train position, rotation and open state classification.
Further details are available in Sec~\ref{sec:suppmat_3dlotus} in the appendix.

\subsection{3D-LOTUS++ policy} 
\label{sec:method_policy++}

To accomplish a task specified by an instruction $L$ like `open the door', the end-to-end policy 3D-LOTUS integrates multiple components into a single action prediction step. This includes task planning (e.g., first grasp the door handle), object grounding (e.g., localize the door handle in 3D space), and motion control (move to the localized door handle).
This integration not only complicates error diagnosis, but also poses challenges in generalization to unseen scenarios, such as `open a new door' or `close the door'.

To alleviate the above limitations, we propose enhancing 3D-LOTUS with foundation models.
Existing LLMs~\cite{llama3modelcard} and VLMs~\cite{kirillov2023sam,radford2021clip,minderer2023owlv2}, are able to generalize to unseen scenarios due to the training on massive data.
Therefore, we introduce a modular framework that disentangles task planning, object grounding and motion control, leveraging the generalization capabilities of foundation models alongside the action execution abilities of 3D-LOTUS to achieve more generalizable robotic manipulation.
Figure~\ref{fig:method_overview} illustrates the overall framework, comprising three modules: task planning with LLMs, object grounding with VLMs, and motion control with a modified version of 3D-LOTUS.

\noindent \textbf{Task planning with LLM.} 
Task planning aims to decompose the instruction $L$ into a sequence of  steps $l_1, \cdots, l_T$.
Each step corresponds to an action primitive that interacts with an object.
In this work, we define six action primitives for object manipulation, covering a broad range of tasks, namely {\it grasp(object)}, {\it move\_grasped\_object(target)}, 
{\it push\_down(object)}, {\it push\_forward(object, target)}, 
{\it release()} and {\it rotate\_grasped\_object()}.
We utilize the LLM LLaMa3-8B~\cite{llama3modelcard} for task planning due to its strong commonsense knowledge and language reasoning capabilities.
By providing prompts for the task requirement and  several in-context examples, we guide the LLM to generate an plan for instruction $L$. Figure~\ref{fig:method_overview} presents an example of generated plans for the task of opening a door.
The detailed prompts for LLMs are presented in Sec~\ref{sec:suppmat_lotus++} in the appendix.

\noindent \textbf{Object grounding with VLMs.} 
This module aims to localize an object given its text description in the generated plan.
To achieve this, we leverage state-of-the-art VLMs to ensure robust generalization to new objects.
First, we employ the open-vocabulary object detector OWLv2~\cite{minderer2023owlv2} to detect bounding boxes with high objectiveness scores for each RGB image. OWLv2 also generates a semantic embedding for each bounding box, which is aligned with text embeddings from the CLIP text encoder~\cite{radford2021clip}.
Next, we use the Segment Anything Model (SAM)~\cite{kirillov2023sam} to segment the object within each bounding box. This segmentation mask, combined with the corresponding RGB-D image, yields a 3D point cloud for each bounding box.
To merge observations of the same object from different cameras, we compare semantic embeddings and point cloud distances. Pairs of objects are merged if their semantic and point cloud distances are below certain thresholds.
In this way, we obtain object-centric representations for all objects in the scene, each object containing a merged point cloud and an averaged semantic embedding.
Given the text description of an object, we compute its text embedding via CLIP and measure cosine similarities between this text embedding and all object semantic embeddings.
The object with the highest cosine similarity is selected as the match.

\noindent \textbf{Motion control.} 
Given the action primitive name and input point cloud, the motion control module predicts a trajectory of actions to execute this actionable step.
3D-LOTUS can be easily modified for this purpose.

First, we change the input point feature $v^o_i$ by leveraging the output of the object grounding module, which segments the manipulated object and/or target location.
This allows us to categorize points into four types: goal object, goal target, robot and obstacle. We treat points that do not belong to goal object, target and robot as obstacles.
We then learn a look-up table to encode each point label and use this as point feature $v^o_i$ instead of RGB colors in addition to the XYZ coordinates $v^p_i$.
The new point feature help the model focus on geometry rather than textures during motion planning, thereby enhancing generalization to objects with novel textures.

Second, instead of predicting a single action, we should generate a sequence of actions to complete the planned step.
To achive this, we introduce a look-up table to encode the timestep index of actions in the trajectory, denoted as $\{x_t\}_{t=1}^{s}$. We concatenate the time embedding $x_t$ with the final point embedding $v_{i}^{e}$ to predict action for each timestep.
The action prediction head is shared across all timesteps.
As the number of actions varies for different plans, we also predict a stop probability to indiciate whether the trajectory should terminate at the current time step.

\section{Experiments}
\label{sec:result}

\subsection{Experimental setup}

\noindent \textbf{Evaluation setup.}
We follow prior work~\cite{shridhar2023peract,chen2023polarnet} and use $K=4$ cameras positioned at the front, left shoulder, right shoulder and wrist, with an image resolution of $256 \times 256$.
For training, we generate 100 demonstrations for each task variation, leading to a dataset of 3,100 demonstrations.
During testing, we use different random seeds from training to ensure initial scene configurations are distinct from the training data.
We evaluate 20 episodes per task variation per seed, and run the evaluation with 5 seeds, which results in $20 \times 5 \times 92$ evaluation episodes in total.
The maximum number of steps per episode is set to 25.
We measure the task performance by success rate (SR) of the evaluation episodes, which is 1 for success and 0 for failure of an episode.
The average SR and standard derivation across seeds are reported.

\noindent \textbf{Implementation details.}
For the 3D-LOTUS model, we use 5 downsampling-upsampling blocks, each containing 1 transformer layer. The hidden sizes for these blocks are 64, 128, 256, 512, 768, respectively. 
For action prediction, the number of bins for position is 30 ($m=15$) and bin size $b=$1cm. The number of bins for rotation is 72 with bin size of 5$^{\circ}$.
In the modified 3D-LOTUS for trajectory prediction, the maximum trajectory length $s$ is set as 5.
We train 3D-LOTUS with batch size of 8 and initial learning rate of 1e-4 for 150k iterations with linear learning rate decay.
Training takes around 11 hours on a single Nvidia A100 GPU.
A validation set of 20 episodes per task variation on Level 1 (different from testing episodes) is used to select the best checkpoint, evaluated every 10k iterations.

\noindent \textbf{Baselines.}
We run four state-of-the-art methods on GemBench, including two 2D image based models\footnote{To be noted, the two methods utilize RGB images as input to the model, but depth images are still used in post-processing to predict actions.} (Hiveformer~\cite{guhur2023hiveformer} and RVT-2~\cite{goyal2024rvt2}), and two 3D-based models (Polarnet~\cite{chen2023polarnet} and 3D diffuser actor~\cite{ke20243ddifusseractor}). 
All the baselines use CLIP text encoder~\cite{radford2021clip}, while only 3D diffuser actor employs visual representations pretrained on large-scale datasets.
We use official codes provided by the authors to validate the training pipeline on RLBench-18Task benchmark. After reproducing the results on RLBench-18Task, we apply the same configuration to train on our GemBench benchmark.

\begin{table}
\centering
\caption{Performance on RLBench-18Task. The Avg. Rank denotes the averaged rank of the model across tasks. Training time is the number of V100 GPU days for training. }
\label{tab:18tasks_sota_cmpr}
\begin{tabular}{lccc} \toprule
 & Avg. SR $\uparrow$ & Avg. Rank $\downarrow$ & Train time $\downarrow$ \\ \midrule
C2F-ARM-BC~\cite{james2022c2farm} & 20.1 & 8.6 & - \\
Hiveformer~\cite{guhur2023hiveformer} & 45.3 & 6.9 & - \\
PolarNet~\cite{chen2023polarnet} & 46.4 & 6.4 & 8.9 \\
PerAct~\cite{shridhar2023peract} & 49.4 & 6.2 & 128.0 \\
RVT~\cite{goyal2023rvt} & 62.9 & 4.4 & 8.0 \\
Act3D~\cite{gervet2023act3d} & 65.0 & 4.3 & 40.0 \\
RVT2~\cite{goyal2024rvt2} & 81.4 & 2.4 & 6.6 \\
3D diffuser actor~\cite{ke20243ddifusseractor} & 81.3 & 2.3 & 67.6 \\ \midrule
3D-LOTUS & \textbf{83.1$_{\pm 0.8}$} & \textbf{2.2} & \textbf{2.2\footnotemark} \\ \bottomrule
\end{tabular}
\end{table}

\begin{table}
\centering
\caption{Performance on four levels of GemBench.}
\label{tab:gembench_sota_cmpr}
\begin{tabular}{lcccc} \toprule
Method & L1 & L2 & L3 & L4 \\ \midrule
Hiveformer~\cite{guhur2023hiveformer} & 60.3$_{\pm 1.5}$ &  26.1$_{\pm 1.4}$ &  35.1$_{\pm 1.7}$ &  0.0$_{\pm 0.0}$ \\
PolarNet~\cite{chen2023polarnet} &  77.7$_{\pm 0.9}$ &  37.1$_{\pm 1.4}$ & 38.5$_{\pm 1.7}$ & 0.1$_{\pm 0.2}$\\
3D diffuser actor~\cite{ke20243ddifusseractor} & 91.9$_{\pm 0.8}$ &  43.4$_{\pm 2.8}$ & 37.0$_{\pm 2.2}$ & 0.0$_{\pm 0.0}$ \\
RVT-2~\cite{goyal2024rvt2} &  89.1$_{\pm 0.8}$ & 51.0$_{\pm 2.3}$ & 36.0$_{\pm 2.2}$ & 0.0$_{\pm 0.0}$ \\ \midrule
3D-LOTUS & \textbf{94.3$_{\pm 1.4}$} & 49.9$_{\pm 2.2}$ & 38.1$_{\pm 1.1}$ & 0.3$_{\pm 0.3}$ \\ 
3D-LOTUS++ & 68.7$_{\pm 0.6}$ & \textbf{64.5$_{\pm 0.9}$} & \textbf{41.5$_{\pm 1.8}$} & \textbf{17.4$_{\pm 0.4}$} \\ \bottomrule
\end{tabular}
\end{table}

\subsection{Comparison with state of the arts}

\noindent \textbf{RLBench-18Task.} 
We first evaluate on the widely used RLBench-18Task benchmark~\cite{shridhar2023peract}, which contains the same 18 tasks (249 variations) for training and testing. %
The results are summarized in Table~\ref{tab:18tasks_sota_cmpr}, with a breakdown of performance on individual tasks provided in Table~\ref{tab:18tasks_sota_cmpr_detail} in the appendix.
The 3D-LOTUS policy achieves state-of-the-art performance using significantly less training time, demonstrating strong action execution capability.

\footnotetext{We use 1 V100 GPU for training for fair comparison of training time.}

\noindent \textbf{GemBench.} 
Table~\ref{tab:gembench_sota_cmpr} presents the results of different models across the four generalization levels in GemBench. Detailed results on individual tasks are shown in Sec~\ref{sec:suppmat_results} in the appendix.
As expected, Level 1 which only involves novel object placements, is easiest. The performance trend of different models are similar to those in RLBench-18Task. 
In Levels 2 to 4, we observe a significant drop in performance for the state-of-the-art methods, highlighting the limitations of existing methods in unseen generalization. 
Generalizing skills for articulated objects (Level 3) proves to be more challenging than for rigid objects (Level 2). 
Level 4, which features long-horizon tasks, is most difficult; the performance of all state-of-the-art methods drop to close to a 0\% success rate.
3D-LOTUS++ significantly outperforms previous methods on more challenging generalization levels.
Note that its performance is lower than the SOTA methods on Level 1. This can be explained by the zero-shot grounding models, which struggles to distinguish some objects in seen tasks such as `tuna can' and `soup can'. 
The performance on Level 4 is suboptimal. Refer to the ablation study for an analysis.

\subsection{Ablations}

\noindent \textbf{3D-LOTUS components.}
In Table~\ref{tab:ablation_polarnet++}, we ablate different components of 3D-LOTUS. 
Row 1 uses regression for action prediction.  Its performance is worse than classification in Row 2 for all levels. Furthermore, it requires more iterations to converge.  
Row 2 and 3 compare two variants for language conditioning. The cross attention outperforms adaptive normalization method, though at a higher computation cost.

\begin{table}
\centering
\tabcolsep=0.12cm
\caption{Ablation of 3D-LOTUS components.}
\label{tab:ablation_polarnet++}
\begin{tabular}{cccccc} \toprule
Action & Condition & L1 & L2 & L3 & L4 \\ \midrule
Regression & AdaptiveNorm & 83.3$_{\pm 0.7}$ & 29.3$_{\pm 1.9}$ & 34.5$_{\pm 1.0}$ & 0.0$_{\pm 0.0}$ \\
Classification & AdaptiveNorm & 90.8$_{\pm 0.7}$ & 47.8$_{\pm 0.6}$ & 37.9$_{\pm 1.5}$ & 0.0$_{\pm 0.0}$ \\ 
Classification & CrossAttn & \textbf{94.3$_{\pm 1.4}$} & \textbf{49.9$_{\pm 2.2}$} & \textbf{38.1$_{\pm 1.1}$} & \textbf{0.3$_{\pm 0.3}$}  \\ \bottomrule
\end{tabular}
\end{table}

\begin{table}[t]
\centering
\caption{Ablations on 3D-LOTUS++ modules.}
\label{tab:ablation_genpolarnet++}
\begin{tabular}{ccccccc} \toprule
\begin{tabular}[c]{@{}c@{}}Task \\ Planning\end{tabular} & \begin{tabular}[c]{@{}c@{}}Object \\ Grounding\end{tabular} & L1 & L2 & L3 & L4 \\ \midrule
GT & GT & \textbf{92.6$_{\pm 0.7}$} & \textbf{80.1$_{\pm 0.5}$} & \textbf{47.8$_{\pm 1.4}$} & \textbf{31.5$_{\pm 1.1}$} \\
GT & VLM & 71.0$_{\pm 1.7}$ & 66.3$_{\pm 0.9}$ & 46.0$_{\pm 1.5}$ & 19.4$_{\pm 1.5}$ \\
LLM & VLM & 68.7$_{\pm 0.6}$ & 64.5$_{\pm 0.9}$ & 41.5$_{\pm 1.8}$ & 17.4$_{\pm 0.4}$ \\ \bottomrule
\end{tabular}
\end{table}

\noindent \textbf{3D-LOTUS++ modules.}
The proposed 3D-LOTUS++ allows for detailed error analysis by isolating each module — task planning, object grounding, and motion control. To facilitate this, we manually annotate groundtruth task plans and object grounding labels for each task, and evaluate the model's performance with and without the ground truth information. The results are shown in Table~\ref{tab:ablation_genpolarnet++}.
We can see that the primary bottleneck in Levels 1 and 2 is the object grounding module, where groundtruth object labels improve the performance by a large margin. In Levels 3 and 4, however, even when provided with all groundtruth information, the performance remains suboptimal. 
The primary issue lies within the 3D-LOTUS motion control policy, specifically its struggle to generalize to long-horizon tasks where initial robot configurations deviate substantially from the training data.
Task planning in Level 4 also suffers from reduced accuracy as the LLM operates without visual input.  This lack of visual awareness can lead to incorrect assumptions about the environment. For example, for the task 'take shoes out of the box,' the LLM cannot determine if the box is open or closed, potentially leading to an inefficient or failed plan."

\begin{figure*}[tp]
    \centering    %
    \includegraphics[width=0.95\linewidth]{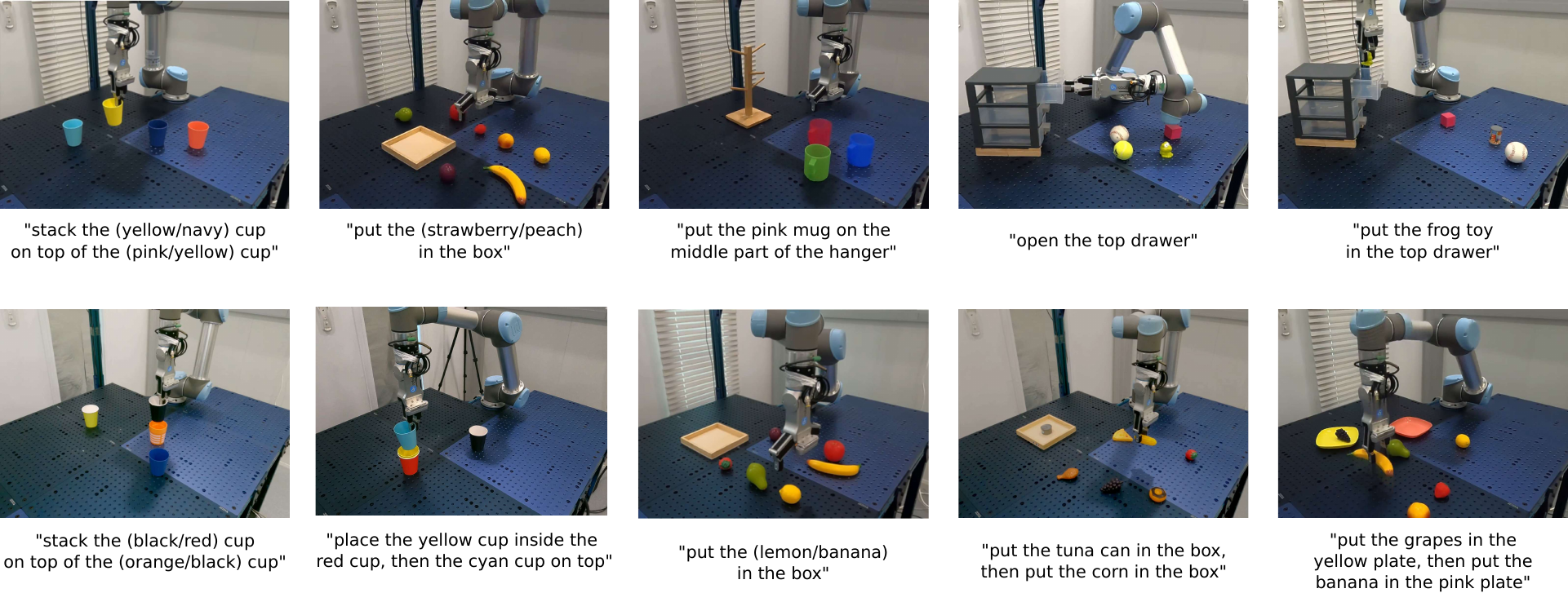}
    \caption{\textbf{Real robot tasks variations.} The top row illustrates task variations used for model training. The bottom row presents new task variations to assess model's generalization capabilities on the real robot.}
    \label{fig:real_robot_tasks}
\end{figure*}

\subsection{Real world experiments}

We further perform real world evaluations of our models.

\noindent \textbf{Experimental setup.}
Our real robot setup includes a 6-DoF UR5 robotic arm equipped with three RealSense d435 cameras. 
We consider 7 variations across 5 tasks during training: stack cup (yellow in pink or navy in yellow), put fruit (strawberry or peach) in box, open drawer, put item in drawer and hang mug. For each task variation, we collect 20 human demonstrations via tele-operation. 
Then, we evaluate on the same 7 seen task variations with different objects placements and evaluate generalization capabilities on 7 new unseen task variations: put fruit (lemon and banana) in box, put food (tuna can then corn) in box and put fruits in plates (grapes in the yellow plate and banana in the pink plate). 
These tasks are illustrated in Figure~\ref{fig:real_robot_tasks}.
For each task variation we run models 10 times and report the success rate. 

Table~\ref{tab:real_robot_seen} shows that our new method, 3D-LOTUS, outperforms PolarNet, achieving an average success rate of $8.1/10$ compared to PolarNet's average performance of $6.7/10$. However, when applying the same 3D-LOTUS model to new task variations, we observe a complete failure to generalize to the new objects and instructions as shown in Table~\ref{tab:real_robot_unseen}. In contrast, our improved model, 3D-LOTUS++, successfully addresses these new task variations, achieving an average success rate of $7.9/10$.

\begin{table}[t]
\centering
\caption{Performance of seen tasks with real robot.}
\label{tab:real_robot_seen}
\begin{tabular}{lcc}
\toprule
\multicolumn{1}{l}{Task} & PolarNet  & 3D-LOTUS  \\ \midrule
Stack yellow cup in pink cup   &    \textbf{10/10}  & 9/10  \\
Stack navy cup in yellow cup   &   9/10   & \textbf{10/10} \\
Put strawberry in box          &   7/10   & \textbf{10/10} \\
Put peach in box               &   \textbf{8/10}   & \textbf{8/10} \\ 
Open drawer                    & 6/10 & \textbf{9/10}   \\
Put item in drawer             & 1/10  & \textbf{3/10}\\
Hang mug                       &  6/10 & \textbf{8/10}  \\
\midrule \midrule
Avg.                           & 6.7/10 & \textbf{8.1/10}  \\\bottomrule
\end{tabular}
\end{table}

\begin{table}
\centering
\caption{Performance of unseen tasks with real robot.}
\label{tab:real_robot_unseen}
\begin{tabular}{lcc}
\toprule
\multicolumn{1}{l}{Task} & 3D-LOTUS  & 3D-LOTUS++  \\ \midrule
Stack red cup in yellow cup & 0/10 & \textbf{8/10} \\
Stack black cup in orange cup & 0/10 & \textbf{7/10} \\
Place the yellow cup inside the red cup, \\ then the cyan cup on top
& 0/10 & \textbf{7/10} \\
Put lemon in box      &   0/10   & \textbf{9/10} \\
Put banana in box     &   0/10   & \textbf{7/10} \\ 
Put tuna can in box, then corn in box  & 0/10 &  \textbf{8/10} \\
Put grapes in yellow plate, \\then banana in pink plate   & 0/10 & \textbf{9/10}  \\
\midrule \midrule
Avg.        & 0/10 & \textbf{7.9/10}  \\\bottomrule
\end{tabular}
\end{table}

\vspace{-0.5em}
\section{Conclusion}
\label{sec:conclusion}

In this work, we introduce a new benchmark and method for generalizable vision-language robotic manipulation.
The proposed benchmark GemBench systematically evaluates four generalization levels: new placements, new rigid objects, new articulated objects, and long-horizon tasks.
To improve generalization ability, we introduce 3D-LOTUS++, a modular framework that leverages foundation models for task planning and object grounding alongside a strong 3D-based motion plan policy 3D-LOTUS.  
Extensive experiments demonstrate the effectiveness of 3D-LOTUS++ on novel tasks.
Our ablation studies highlight object grounding as a critical bottleneck and reveal the limitations of the motion control policy in complex scenarios. Future work will focus on addressing these two issues.

\noindent\textbf{Acknowledgements.} This work was partially supported by the HPC resources from GENCI-IDRIS (Grant 20XX-AD011012122 and AD011014846). 
It was funded in part by the French government under management of Agence Nationale de la Recherche as part of the “France 2030" program, reference ANR-23-IACL-0008 (PR[AI]RIE-PSAI projet), and the ANR project VideoPredict (ANR-21-FAI1-0002-01). Cordelia Schmid would like to acknowledge the support by the Körber European Science Prize.


\section*{APPENDIX}

\subsection{The proposed GemBench benchmark}
\label{sec:suppmat_gembench}

Table~\ref{tab:gembench_tasks} presents all the tasks and variations used in training and the four generalization levels in testing in the proposed GemBench.

\begin{table*}[t]
\centering
\tabcolsep=0.18cm
\caption{\textbf{Training and testing tasks \& variations in GemBench.} The testing tasks contain four levels of generalization, where Level 1 evaluates the generalization to novel placements, Level 2 novel rigid objects, Level 3 novel articulated objects, and Level 4 novel long-horizon tasks.}
\label{tab:gembench_tasks}
\begin{tabular}{llccccccc}
\toprule
\rowcolor[HTML]{CBCEFB} 
\cellcolor[HTML]{CBCEFB} & \multicolumn{2}{c}{\cellcolor[HTML]{CBCEFB}Train / Level 1} & \multicolumn{2}{c}{\cellcolor[HTML]{CBCEFB}Level 2} & \multicolumn{3}{c}{\cellcolor[HTML]{CBCEFB}Level 3} & Level 4 \\
\rowcolor[HTML]{CBCEFB} 
\multirow{-2}{*}{\cellcolor[HTML]{CBCEFB}} & Task & Variation & Color & Shape & Instance & Category & Action-Part & Long-horizon \\ \midrule
 &  & maroon button & azure button &  &  &  &  & 2 buttons \\
 &  & navy button & rose button &  &  &  &  & 3 buttons \\
\multirow{-3}{*}{Press} & \multirow{-3}{*}{Push button} & yellow button & white button & \multirow{-3}{*}{Lamp on} &  &  &  & 4 buttons \\ \midrule
\rowcolor[HTML]{EFEFEF} 
\cellcolor[HTML]{EFEFEF} & \cellcolor[HTML]{EFEFEF} & red block & teal block & red cylinder &  &  &  &  \\
\rowcolor[HTML]{EFEFEF} 
\cellcolor[HTML]{EFEFEF} & \cellcolor[HTML]{EFEFEF} & lime block & violet block & red star &  &  &  &  \\
\rowcolor[HTML]{EFEFEF} 
\cellcolor[HTML]{EFEFEF} & \multirow{-3}{*}{\cellcolor[HTML]{EFEFEF}Pick and lift} & cyan block & black block & red moon &  &  &  &  \\ %
\rowcolor[HTML]{EFEFEF} 
\cellcolor[HTML]{EFEFEF} & \cellcolor[HTML]{EFEFEF} & magenta cup & gray cup & \cellcolor[HTML]{EFEFEF} &  &  &  &  \\
\rowcolor[HTML]{EFEFEF} 
\cellcolor[HTML]{EFEFEF} & \cellcolor[HTML]{EFEFEF} & silver cup & olive cup & \cellcolor[HTML]{EFEFEF} &  &  &  &  \\
\rowcolor[HTML]{EFEFEF} 
\multirow{-6}{*}{\cellcolor[HTML]{EFEFEF}Pick} & \multirow{-3}{*}{\cellcolor[HTML]{EFEFEF}Pick up cup} & orange cup & purple cup & \multirow{-3}{*}{\cellcolor[HTML]{EFEFEF}red toy} &  &  &  &  \\ \midrule
 &  & green target & pink target &  &  &  &  &  \\
 & \multirow{-2}{*}{Slide block} & blue target & yellow target &  &  &  &  &  \\ %
 &  & teal target & cyan target &  &  &  &  &  \\ 
\multirow{-4}{*}{Push} & \multirow{-2}{*}{Reach and drag} & black target & navy target &  &  &  &  &  \\ \midrule
\rowcolor[HTML]{EFEFEF} 
\cellcolor[HTML]{EFEFEF} & \cellcolor[HTML]{EFEFEF} & azure jar & blue jar &  &  &  &  &  \\
\rowcolor[HTML]{EFEFEF} 
\cellcolor[HTML]{EFEFEF} & \multirow{-2}{*}{\cellcolor[HTML]{EFEFEF}Close jar} & violet jar & green jar &  &  &  &  &  \\ %
\rowcolor[HTML]{EFEFEF} 
\cellcolor[HTML]{EFEFEF} & \cellcolor[HTML]{EFEFEF} & rose bulb & lime bulb &  &  &  &  &  \\
\rowcolor[HTML]{EFEFEF} 
\multirow{-4}{*}{\cellcolor[HTML]{EFEFEF}Screw} & \multirow{-2}{*}{\cellcolor[HTML]{EFEFEF}Screw bulb} & white bulb & maroon bulb &  &  &  &  &  \\ \midrule
 & Close fridge & fridge &  &  & fridge2 &  & door &  \\
 & Close laptop lid & laptop lid &  &  & laptop lid2 &  & box &  \\
\multirow{-3}{*}{Close} & Close microwave & microwave &  &  & microwave2 & \multirow{-3}{*}{grill} & drawer &  \\ \midrule
\rowcolor[HTML]{EFEFEF} 
\cellcolor[HTML]{EFEFEF} & Open door & door &  &  & door2 & \cellcolor[HTML]{EFEFEF} & fridge & \cellcolor[HTML]{EFEFEF} \\
\rowcolor[HTML]{EFEFEF} 
\cellcolor[HTML]{EFEFEF} & Open box & box &  &  & box2 & \multirow{-1}{*}{\cellcolor[HTML]{EFEFEF}toilet} & laptop lid & \multirow{-2}{*}{\cellcolor[HTML]{EFEFEF}\begin{tabular}[c]{@{}c@{}}Take shoes \\ out of box\end{tabular}} \\
\rowcolor[HTML]{EFEFEF} 
\cellcolor[HTML]{EFEFEF} & \cellcolor[HTML]{EFEFEF} & bottom drawer &  &  & drawer2, drawer3 & \cellcolor[HTML]{EFEFEF} & microwave & \cellcolor[HTML]{EFEFEF} \\
\rowcolor[HTML]{EFEFEF} 
\multirow{-4}{*}{\cellcolor[HTML]{EFEFEF}Open} & \multirow{-2}{*}{\cellcolor[HTML]{EFEFEF}Open drawer} & top drawer &  &  & long drawer w/ 4 levels &  & middle drawer & \multirow{-2}{*}{\cellcolor[HTML]{EFEFEF}\begin{tabular}[c]{@{}c@{}}Put 3 items \\ in drawer\end{tabular}} \\ \midrule
 &  & 2 gray blocks & 2 orange blocks &  &  &  &  &  \\
 &  & 2 olive blocks & 2 silver blocks &  &  &  &  & \multirow{-2}{*}{\begin{tabular}[c]{@{}c@{}}Stack 3-4 \\ blocks\end{tabular}} \\
 & \multirow{-3}{*}{Stack blocks} & 2 purple blocks & 2 magenta blocks &  &  &  &  & Stack 2 cups \\ \cmidrule{2-9}
 &  & crackers box &  & mustard bottle &  &  &  &  \\
 & \multirow{-2}{*}{Put groceries} & soup can &  & sugar box &  &  &  & \multirow{-2}{*}{\begin{tabular}[c]{@{}c@{}}Put all \\ groceries\end{tabular}} \\ \cmidrule{2-9} 
 &  & bottom shelf &  &  &  &  &  &  \\
\multirow{-7}{*}{\begin{tabular}[c]{@{}c@{}}Put/\\ Stack\end{tabular}} & \multirow{-2}{*}{Put money} & middle shelf &  & \multirow{-2}{*}{\begin{tabular}[c]{@{}c@{}}Put cube in \\ bottom shelf\end{tabular}} &  &  & \multirow{-2}{*}{top shelf} &  \\ \bottomrule
\end{tabular}
\end{table*}

\subsection{Details of 3D-LOTUS policy}
\label{sec:suppmat_3dlotus}

\subsubsection{Point cloud preprocessing}
\begin{figure*}
     \centering
     \begin{subfigure}[b]{0.32\textwidth}
         \centering
         \includegraphics[width=\textwidth]{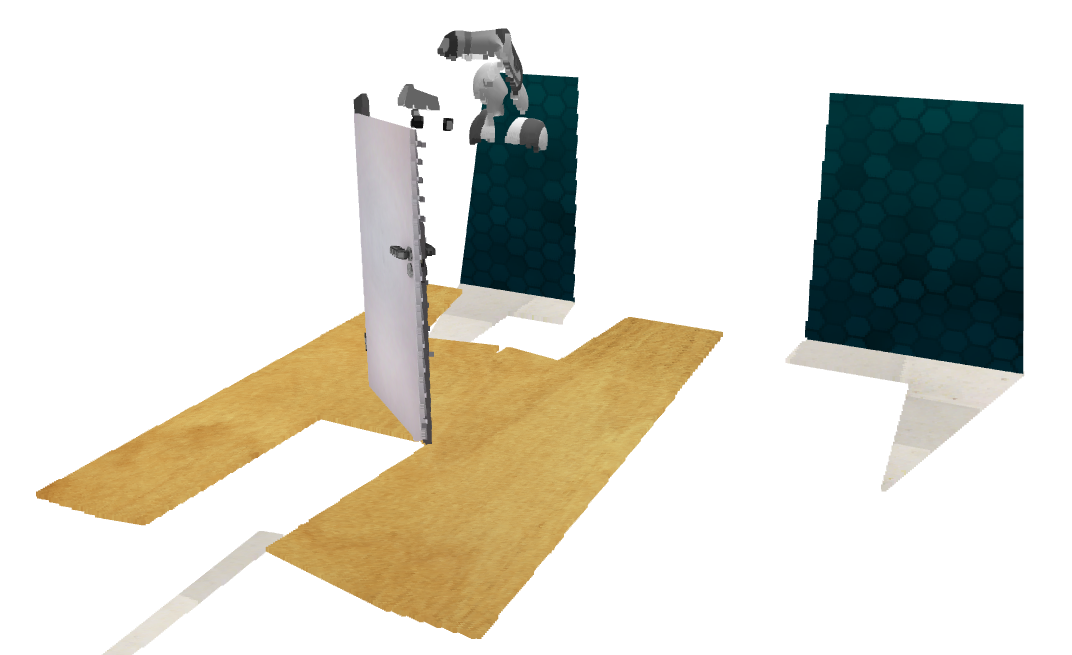}
         \caption{The original point cloud.}
         \label{fig:point_cloud_v1}
     \end{subfigure}
     \hfill
     \begin{subfigure}[b]{0.32\textwidth}
         \centering
         \includegraphics[width=\textwidth]{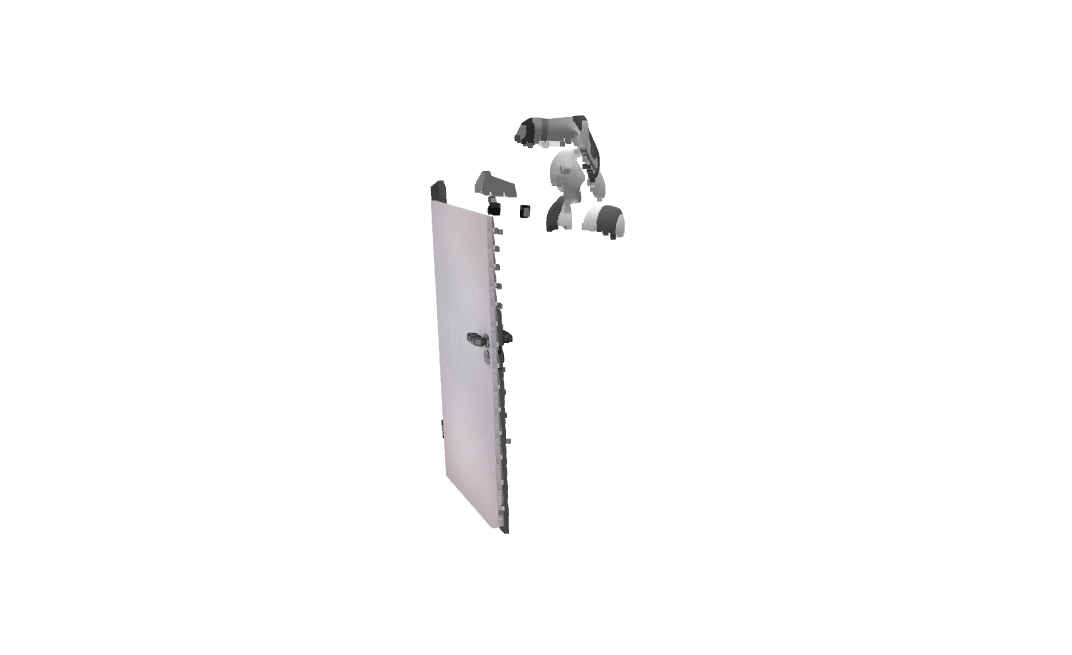}
         \caption{Removing background and table.}
         \label{fig:point_cloud_v2}
     \end{subfigure}
     \hfill
     \begin{subfigure}[b]{0.32\textwidth}
         \centering
         \includegraphics[width=\textwidth]{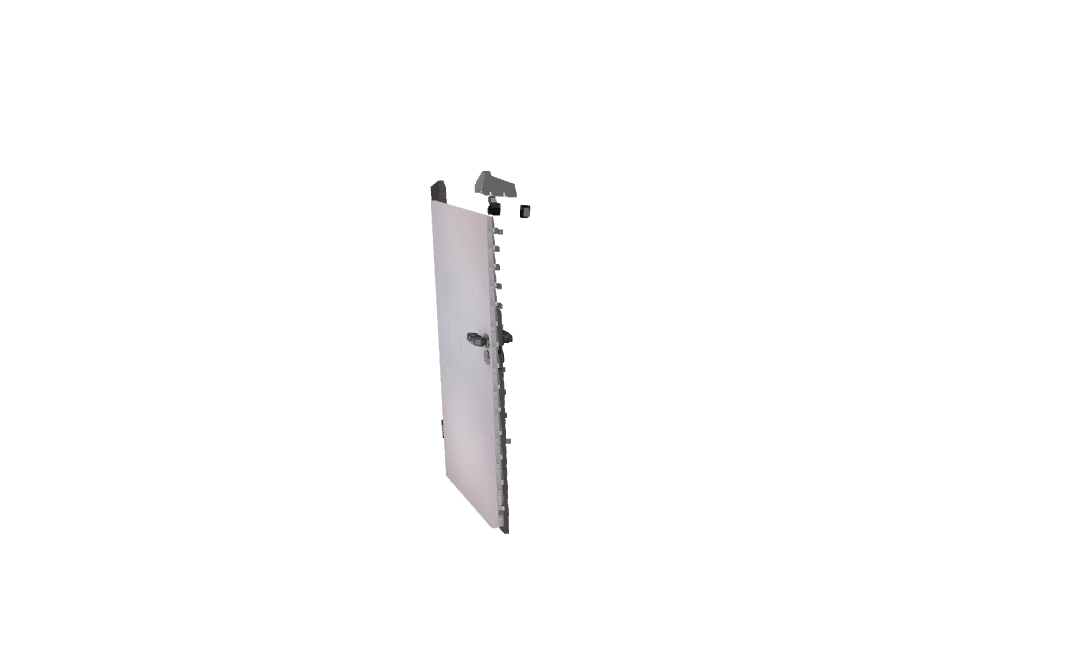}
         \caption{Removing robot arm.}
         \label{fig:point_cloud_v3}
     \end{subfigure}
        \caption{\textbf{Automatic point removal.} We use geometry information to automatically filter out irrelevant points in the scene.}
        \label{fig:point_removal}
\end{figure*}

We automatically filter out irrelevant points from the  point cloud during preprocessing.
To remove background and table points, we define the robot's workspace and the table height, excluding all points outside these boundaries.
For the robotic arm, we assign 3D bounding boxes to each of its links. Using the robot's proprioceptive state, we transform these bounding boxes based on the known poses and remove any points within them.
As a result, the remaining point cloud contains only  objects and the robotic gripper.
Figure~\ref{fig:point_removal} illustrates the point cloud before and after the point removal step. 

\subsubsection{Action prediction}
We discretize the groundtruth position $a^p_t$ and rotation $a^r_t$ to train the model.
For position prediction, let $b^p_{t,k} \in \mathbb{R}^{n \times 2m}$ represent the position of the concatenated bins for all points along the $k$ axis, where $n$ is the number of points and $2m$ is the number of bins per point. 
We calculate the Euclidean distance between each bin and $a^p_{t,k}$, defining the score for bin $i$ as:
\begin{equation}
    \hat{p}_{t,k,i} = \left\{
    \begin{aligned}
         & 0 , ~\text{if}~ || b^p_{t,k,i} - a^p_{t,k}||_2^2 > 0.01 ~\text{or}~ b^p_{t,k,i} \in \mathbb{B} \\
        & \frac{1}{|| b^p_{t,k,i} - a^p_{t,k}||_2^2}, ~\text{otherwise.}
    \end{aligned}
    \right.
\end{equation}
where $\mathbb{B}$ denotes the set of points that belong to the robot arm and gripper, and their scores are set to zero to ensure that the gripper's position is predicted based only on the objects in the scene.
The groundtruth probability of position along the $k$ axis is then obtained by normalizing the scores $\hat{p}_{t,k,i}$ via L1 norm.
For rotation prediction, we simply use one-hot label along each axis.

\subsection{Details of 3D-LOTUS++ policy}
\label{sec:suppmat_lotus++}

\subsubsection{Task planning}
Figure~\ref{fig:llm_prompts} illustrates the prompts used in LLMs for task planning.
For each task variation in the training set, we craft a corresponding example as shown in Figure~\ref{fig:llm_incontext_examples}.
During inference, for each new instruction, we use SentenceBert~\cite{reimers2019sentencebert} to compute the sentence embedding of the instruction and compare it to the existing instructions. We then select the top 20 examples with the highest similarities to the query instruction as in-context examples to the LLM.

\begin{figure*}
\begin{lstlisting}[language=Python,style=style-prompt]
I would like you to help me write Python code to control a robot arm operating in a tabletop environment. Please complete the code every time when I give you new query and a list of objects visible at the initial step. Pay attention to appeared patterns in the given context code. Be thorough and thoughtful in your code. Do not include any import statement. Do not repeat my question. Do not provide any text explanation (comment in code is okay).

You are only allowd to use the following action primitives that a robotic arm can perform:

1. `grasp(object)`: Grasp the specified object. Ensure that the robot gripper is open and not holding any other object before grasping. The robot gripper can only grasp one object at a time. After grasping, the robot gripper will close and securely hold the object. Return the grasped object.

2. `move_grasped_object(target)`: Move the grasped object to the specified target. Ensure that the robot gripper is closed and holding an object before moving. After moving, the robot gripper will still hold the object. The target can be a text description of a specified place, the location of previous objects, or a direction such as up, down, forward and out for small movements in those directions. Return the grasped object.

3. `rotate_grasped_object()`: Rotate the gripper while holding the object. Ensure that the robot gripper is holding an object before performing the rotation. After rotating, the gripper will still hold the object. Return the grasped object.

4. `push_down(object)`: Push down the specified object vertically, such as a button. The robot gripper does not hold the specified object but may hold other objects. Return the grasped object.

5. `push_forward(object, target)`: Push forward the specified object towards a target place. If no target is specified, the object will be pushed forward by a small distance.The robot gripper does not hold the specified object but may hold other objects. Return the grasped object.

6. `release()`: Open the gripper to release an object. Ensure the object is held in the gripper before releasing. After releasing, the gripper is open and not holding any object.

It's essential to stick to the format of these basic skills. When creating a plan, replace object or target inside the function with text descriptions or previously returned objects. Do not use objects not visible in the scene, but the robot can discover more objects through for example openning box or drawer. Generate step-by-step plans. Do not use for loop.

I will first give you the context of the code below:
\end{lstlisting}
\caption{\textbf{Prompts used in LLM for task planning.}}
\label{fig:llm_prompts}
\end{figure*}

\begin{figure*}
\begin{lstlisting}[language=Python,style=style-prompt]
# query: push the maroon button.
button = push_down(object="maroon button")

# query: close fridge.
fridge_door = push_forward(object="fridge door")

# query: close laptop lid.
laptop_lid = grasp(object="laptop lid")
laptop_lid = move_grasped_object(target="down")
release()

# query: close microwave.
microwave_door = push_forward(object="microwave door")

# query: open the door.
door_handle = grasp(object="door handle")
door_handle = rotate_grasped_object()
door_handle = push_forward(object=door_handle)

# query: open box.
box_lid = grasp(object="box lid")
box_lid = move_grasped_object(target="up")
release()

# query: open bottom drawer.
bottom_handle = grasp(object="bottom drawer handle")
bottom_handle = move_grasped_object(target="out")
release()

# query: lift the cyan block up to the target.
cyan_cube = grasp(object="cyan cube")
cyan_cube = move_grasped_object(target="red ball")

# query: lift the orange cup.
orange_cup = grasp(object="orange cup")
orange_cup = move_grasped_object(target="up")

# query: pick up and set down 2 purple blocks on top of each other.
purple_cube_1 = grasp(object="purple cube")
purple_cube_1 = move_grasped_object(target="green square")
release()
purple_cube_2 = grasp(object="purple cube", not=[purple_cube_1])
purple_cube_2 = move_grasped_object(target=purple_cube_1)
release()

# query: put the crackers box in the cupboard.
crackers_box = grasp(object="crakers box")
crackers_box = move_grasped_object(target="cupboard")
release()

# query: leave the money on the middle shelf on the safe.
money = grasp(object="money")
money = move_grasped_object(target="middle shelf")
release()

# query: push the block until it is sitting on top of the green target.
cube = push_forward(object="red cube", target="green square")

# query: use the stick to drag the cube onto the teal target.
stick = grasp(object="stick")
cube = push_forward(object="gray cube", target="teal square")

# query: screw on the violet jar lid.
lid = grasp(object="gray lid")
lid = move_grasped_object(target="violet jar")
lid = rotate_grasped_object()
release()

# query: screw in the white light bulb.
bulb = grasp(object="white light bulb")
bulb = move_grasped_object(target="brown lamp")
bulb = rotate_grasped_object()
release()
\end{lstlisting}
\caption{\textbf{In-context examples of each training task for task planning.}}
\label{fig:llm_incontext_examples}
\end{figure*}

\begin{figure*}
\begin{lstlisting}[language=Python,style=style-prompt]
Please help me define the height range for different levels of an articulated object. I will provide you the target level and the total height of the object. Your task is to output two numbers representing the height range for the target level. Pay attention to the appeared patterns in the given examples. Do not repeat my question. Do not provide any text explanation. 

The examples are as follows: 

target: bottom drawer handle
height: 0.4
target height range: [0.1, 0.2]

target: top drawer handle
height: 0.4
target height range: [0.3, 0.4]

target: bottom shelf
height: 0.5
target height range: [0, 0.1]

target: middle shelf
height: 0.5
target height range: [0.15, 0.25]
\end{lstlisting}
\caption{\textbf{Prompts used in LLMs to predict the height range of an object.}}
\label{fig:llm_height_prompt}
\end{figure*}

\subsubsection{Object grounding}
We use VLMs to detect the location of queried objects except for tasks requiring grounding different parts of articulated object like bottom drawer and top shelf.
This limitation arises because our VLMs can only detect the whole object like drawer, but cannot ground the target level of the drawer.
Therefore, we further leverage the LLM to predict the height range of the target object. 
We first obtain the overall height of the target object based on VLM's prediction, then we use prompts presented in Figure~\ref{fig:llm_height_prompt} to guide the LLM in predicting the height range.

\begin{table*}[t]
\centering
\Huge
\caption{\textbf{Multi-Task Performance on RLBench.} We report the success rate on all tasks in RLBench-18Task~\cite{shridhar2023peract} benchmark. Our 3D-LOTUS outperforms all methods while having higher training speed.}
\label{tab:18tasks_sota_cmpr_detail}
\resizebox{\textwidth}{!}{
\begin{tabular}{lcccccccccccccccccc} 
\toprule
\rowcolor[HTML]{CBCEFB}
& Avg. & Avg. & Train time &  Inf. Speed & Close & Drag & Insert  & Meat off & Open & Place & Place \\
\rowcolor[HTML]{CBCEFB}
Models & Success $\uparrow$ & Rank $\downarrow$ & (in days) $\downarrow$ &  (in fps) $\uparrow$ & Jar & Stick & Peg & Grill & Drawer & Cups & Wine  \\ 
\midrule
C2F-ARM-BC~\cite{james2022c2farm,shridhar2023peract} & 20.1 & 8.6 & - & - & 24 & 24 & 4 & 20 & 20 & 0 & 8 \\
\rowcolor[HTML]{EFEFEF}
HiveFormer~\cite{guhur2023hiveformer} & 45.3 & 6.9 & - & - & 52.0 & 76.0 & 0.0 & \textbf{100.0} & 52.0 & 0.0 & 80 \\
PolarNet~\cite{chen2023polarnet} & 46.4 & 6.4 & 8.9 & - & 36.0 & 92.0 & 4.0 & \textbf{100.0} & 84.0 & 0.0 & 40 \\
\rowcolor[HTML]{EFEFEF}
PerAct~\cite{shridhar2023peract} & 49.4 & 6.2 & 128.0 & 4.9 & 55.2$_{\pm 4.7}$ & 89.6$_{\pm 4.1}$ & 5.6$_{\pm 4.1}$ & 70.4$_{\pm 2.0}$ & 88.0$_{\pm 5.7}$ & 2.4$_{\pm 3.2}$ & 44.8$_{\pm 7.8}$ \\
\rowcolor[HTML]{EFEFEF}
RVT~\cite{goyal2023rvt} & 62.9 & 4.4 & 8.0 & 11.6 & 52.0$_{\pm 2.5}$ & 99.2$_{\pm 1.6}$ & 11.2$_{\pm 3.0}$ & 88.0$_{\pm 2.5}$ & 71.2$_{\pm 6.9}$ & 4.0$_{\pm 2.5}$ & 91.0$_{\pm 5.2}$ \\
Act3D~\cite{gervet2023act3d} & 65.0 & 4.3 & 40.0 & - & 92.0 & 92.0 & 27.0 & 94.0 & \textbf{93.0} & 3.0 & 80 \\
RVT-2~\cite{goyal2024rvt2} & 81.4 & 2.4 & 6.6 & 20.6 & \textbf{100.0$_{\pm 0.0}$} & 99.0$_{\pm 1.7}$ & 40.0$_{\pm 0.0}$ & 99.0$_{\pm 1.7}$ & 74.0$_{\pm 11.8}$ & 38.0$_{\pm 4.5}$ & \textbf{95.0$_{\pm 3.3}$} \\
\rowcolor[HTML]{EFEFEF}
3D diffuser actor~\cite{ke20243ddifusseractor} & 81.3 &  2.3 & 67.6 & - & 96.0$_{\pm 2.5}$ & \textbf{100.0$_{\pm 0.0}$} & 65.6$_{\pm 4.1}$ & 96.8$_{\pm 1.6}$ & 89.6$_{\pm 4.1}$ & 24.0$_{\pm 7.6}$ & 93.6$_{\pm 4.8}$ \\
3D-LOTUS (ours) & \textbf{83.1} & \textbf{2.2} & \textbf{2.2} & 9.5 & 96.0$_{\pm 0.0}$ & \textbf{100.0}$_{\pm 0.0}$ & \textbf{69.6$_{\pm 3.6}$} & 98.4$_{\pm 2.2}$ & 85.6$_{\pm 7.3}$ & \textbf{40.8$_{\pm 12.1}$} & 91.2$_{\pm 6.6}$ \\
\midrule
\rowcolor[HTML]{CBCEFB}
& Push & Put in & Put in & Put in & Screw & Slide & Sort & Stack & Stack & Sweep to & Turn \\
\rowcolor[HTML]{CBCEFB}
Models & Buttons & Cupboard & Drawer & Safe & Bulb & Block & Shape & Blocks & Cups & Dustpan & Tap \\
\midrule
C2F-ARM-BC~\cite{james2022c2farm,shridhar2023peract} & 72 & 0 & 4 & 12 & 8 & 16 & 8 & 0 & 0 & 0 & 68 \\
\rowcolor[HTML]{EFEFEF}
HiveFormer~\cite{guhur2023hiveformer} & 84 & 32.0 & 68.0 & 76.0 & 8.0 & 64.0 & 8.0 & 8.0 & 0.0 & 28.0 & 80  \\
PolarNet~\cite{chen2023polarnet} & 96 & 12.0 & 32.0 & 84.0 & 44.0 & 56.0 & 12.0 & 4.0 & 8.0 & 52.0 & 80  \\
\rowcolor[HTML]{EFEFEF}
PerAct~\cite{shridhar2023peract} & 92.8$_{\pm 3.0}$       & 28.0$_{\pm 4.4}$ & 51.2$_{\pm 4.7}$ & 84.0$_{\pm 3.6}$ & 17.6$_{\pm 2.0}$ & 74.0$_{\pm 13.0}$ & 16.8$_{\pm 4.7}$ & 26.4$_{\pm 3.2}$ & 2.4$_{\pm 2.0}$ & 52.0$_{\pm 0.0}$ & 88.0$_{\pm 4.4}$ \\
Act3D~\cite{gervet2023act3d} & 99 & 51.0 & 90.0 & 95.0 & 47.0 & 93.0 & 8.0 & 12.0 & 9.0 & 92.0 & 94 \\
\rowcolor[HTML]{EFEFEF}
RVT~\cite{goyal2023rvt} & \textbf{100.0$_{\pm 0.0}$} & 49.6$_{\pm 3.2}$ & 88.0$_{\pm 5.7}$ & 91.2$_{\pm 3.0}$ & 48.0$_{\pm 5.7}$ & 81.6$_{\pm 5.4}$ & 36.0$_{\pm 2.5}$ & 28.8$_{\pm 3.9}$ & 26.4$_{\pm 8.2}$ & 72.0$_{\pm 0.0}$ & 93.6$_{\pm 4.1}$ \\
RVT-2~\cite{goyal2024rvt2} & \textbf{100.0$_{\pm 0.0}$} & 66.0$_{\pm 4.5}$ & 96.0$_{\pm 0.0}$ & 96.0$_{\pm 2.8}$  & 88.0$_{\pm 4.9}$ & 92.0$_{\pm 2.8}$ & 35.0$_{\pm 7.1}$ & \textbf{80.0$_{\pm 2.8}$} & 69.0$_{\pm 5.9}$ & \textbf{100.0$_{\pm 0.0}$} & 99.0$_{\pm 1.7}$ \\
\rowcolor[HTML]{EFEFEF}
3D diffuser actor~\cite{ke20243ddifusseractor} & 98.4$_{\pm 2.0}$ & \textbf{85.6$_{\pm 4.1}$} & 96.0$_{\pm 3.6}$ & \textbf{97.6$_{\pm 2.0}$} & 82.4$_{\pm 2.0}$ & 97.6$_{\pm 3.2}$ & \textbf{44.0$_{\pm 4.4}$} & 68.3$_{\pm 3.3}$ & 47.2$_{\pm 8.5}$ & 84.0$_{\pm 4.4}$ & \textbf{99.2$_{\pm 1.6}$} \\
3D-LOTUS (ours) & \textbf{100.0$_{\pm 0.0}$} & 78.4$_{\pm 4.6}$ & \textbf{97.6$_{\pm 3.6}$} & 95.2$_{\pm 3.4}$ & \textbf{88.8}$_{\pm 3.4}$ & \textbf{99.2$_{\pm 1.8}$} & 34.4$_{\pm 4.6}$ & 58.4$_{\pm 8.3}$ & \textbf{75.2$_{\pm 7.7}$} & 96.0$_{\pm 2.8}$ & 90.4$_{\pm 4.6}$ \\
\bottomrule
\end{tabular}
}
\end{table*}

\subsection{Detailed evaluation results}
\label{sec:suppmat_results}

\noindent \textbf{RLBench-18Task.} 
Table~\ref{tab:18tasks_sota_cmpr_detail} presents the detailed results of different models on each task in RLBench-18Task benchmark.
Our 3D-LOTUS policy achieves better performances especially on tasks requiring high precision such as insert peg, place cups and stack cups.

\noindent \textbf{GemBench.}
Table~\ref{tab:gembench_sota_cmpr_l1_detail} to~\ref{tab:gembench_sota_cmpr_l4_detail} show the results of different models on the four generalization levels in GemBench respectively.

\begin{table*}[t]
\tabcolsep=0.08cm
\caption{Performance on GemBench Level 1.}
\label{tab:gembench_sota_cmpr_l1_detail}
\begin{tabular}{lccccccccccc} \toprule
\rowcolor[HTML]{CBCEFB}
Method & Avg. & \begin{tabular}[c]{@{}c@{}}Close\\ Fridge+0\end{tabular} & \begin{tabular}[c]{@{}c@{}}Close\\ Jar+15\end{tabular} & \begin{tabular}[c]{@{}c@{}}Close\\ Jar+16\end{tabular} & \begin{tabular}[c]{@{}c@{}}CloseLaptop\\ Lid+0\end{tabular} & \begin{tabular}[c]{@{}c@{}}Close\\ Microwave+0\end{tabular} & \begin{tabular}[c]{@{}c@{}}LightBulb\\ In+17\end{tabular} & \begin{tabular}[c]{@{}c@{}}LightBulb\\ In+19\end{tabular} & \begin{tabular}[c]{@{}c@{}}Open\\ Box+0\end{tabular} & \begin{tabular}[c]{@{}c@{}}Open\\ Door+0\end{tabular} & \begin{tabular}[c]{@{}c@{}}Open\\ Drawer+0\end{tabular} \\ \midrule
Hiveformer~\cite{guhur2023hiveformer} & 60.3$_{\pm 1.5}$ & 96$_{\pm 4.2}$ & 64$_{\pm 13.9}$ & 92$_{\pm 2.7}$ & 90$_{\pm 3.5}$ & 88$_{\pm 7.6}$ & 12$_{\pm 4.5}$ & 13$_{\pm 6.7}$ & 4$_{\pm 4.2}$ & 53$_{\pm 15.2}$ & 15$_{\pm 12.2}$ \\
\rowcolor[HTML]{EFEFEF}
PolarNet~\cite{chen2023polarnet} & 77.6$_{\pm 0.9}$ & 99$_{\pm 2.2}$ & 99$_{\pm 2.2}$ & 99$_{\pm 2.2}$ & 95$_{\pm 3.5}$ & 98$_{\pm 2.7}$ & 72$_{\pm 12.5}$ & 71$_{\pm 6.5}$ & 32$_{\pm 11.5}$ & 69$_{\pm 8.9}$ & 61$_{\pm 12.4}$ \\
3D diffuser actor~\cite{ke20243ddifusseractor}  & 91.9$_{\pm 0.8}$ & \textbf{100}$_{\pm 0.0}$ & \textbf{100}$_{\pm 0.0}$ & \textbf{100}$_{\pm 0.0}$ & \textbf{99}$_{\pm 2.2}$ & \textbf{100}$_{\pm 0.0}$ & 85$_{\pm 5.0}$ & 88$_{\pm 2.7}$ & 11$_{\pm 2.2}$ & 96$_{\pm 4.2}$ & 82$_{\pm 9.1}$ \\
\rowcolor[HTML]{EFEFEF}
RVT-2~\cite{goyal2024rvt2}  & 89.0$_{\pm 0.8}$ & 77$_{\pm 11.0}$ & 97$_{\pm 4.5}$ & 98$_{\pm 2.7}$ & 77$_{\pm 13.0}$ & \textbf{100}$_{\pm 0.0}$ & \textbf{93}$_{\pm 5.7}$ & \textbf{91}$_{\pm 8.2}$ & 7$_{\pm 4.5}$ & \textbf{98}$_{\pm 4.5}$ & \textbf{93}$_{\pm 5.7}$ \\
\rowcolor[HTML]{EFEFEF}
3D-LOTUS (ours) & \textbf{94.3}$_{\pm 3.5}$ & 96$_{\pm 3.7}$ & \textbf{100}$_{\pm 0.0}$ & \textbf{100}$_{\pm 0.0}$ & 98$_{\pm 2.5}$ & 98$_{\pm 4.0}$ & 84$_{\pm 7.4}$ & 85$_{\pm 9.5}$ & \textbf{99}$_{\pm 2.0}$ & 77$_{\pm 2.5}$ & 83$_{\pm 8.7}$ \\ 
3D-LOTUS++ (ours) & 68.7$_{\pm 0.6}$ & 95$_{\pm 0.0} $ & \textbf{100$_{\pm 0.0}$} & 99$_{\pm 2.0}$ & 28$_{\pm 2.5}$ & 87$_{\pm 5.1}$ & 55$_{\pm 10.5}$ & 45$_{\pm 8.9}$ & 55$_{\pm 8.9}$ & 79$_{\pm 9.7}$ & 68$_{\pm 12.5}$ \\
\midrule

\rowcolor[HTML]{CBCEFB}
Method & \begin{tabular}[c]{@{}c@{}}Open\\ Drawer+2\end{tabular} & \begin{tabular}[c]{@{}c@{}}Pick\&\\ Lift+0\end{tabular} & \begin{tabular}[c]{@{}c@{}}Pick\&\\ Lift+2\end{tabular} & \begin{tabular}[c]{@{}c@{}}Pick\&\\ Lift+7\end{tabular} & \begin{tabular}[c]{@{}c@{}}PickUp\\ Cup+8\end{tabular} & \begin{tabular}[c]{@{}c@{}}PickUp\\ Cup+9\end{tabular} & \begin{tabular}[c]{@{}c@{}}PickUp\\ Cup+11\end{tabular} & \begin{tabular}[c]{@{}c@{}}Push\\ Button+0\end{tabular} & \begin{tabular}[c]{@{}c@{}}Push\\ Button+3\end{tabular} & \begin{tabular}[c]{@{}c@{}}Push\\ Button+4\end{tabular} & \begin{tabular}[c]{@{}c@{}}PutIn\\ Cupboard+0\end{tabular} \\ \midrule
Hiveformer~\cite{guhur2023hiveformer}  & 59$_{\pm 7.4}$ & 86$_{\pm 4.2}$ & 92$_{\pm 6.7}$ & 93$_{\pm 2.7}$ & 83$_{\pm 7.6}$ & 69$_{\pm 12.9}$ & 61$_{\pm 19.8}$ & 84$_{\pm 11.9}$ & 68$_{\pm 6.7}$ & 87$_{\pm 7.6}$ & 34$_{\pm 8.2}$ \\
\rowcolor[HTML]{EFEFEF}
PolarNet~\cite{chen2023polarnet} & 90$_{\pm 7.1}$ & 92$_{\pm 9.1}$ & 84$_{\pm 7.4}$ & 88$_{\pm 5.7}$ & 82$_{\pm 7.6}$ & 79$_{\pm 4.2}$ & 72$_{\pm 10.4}$ & \textbf{100}$_{\pm 0.0}$ & \textbf{100}$_{\pm 0.0}$ & 99$_{\pm 2.2}$ & 52$_{\pm 7.6}$ \\
3D diffuser actor~\cite{ke20243ddifusseractor}  & \textbf{97}$_{\pm 4.5}$ & \textbf{99}$_{\pm 2.2}$ & 99$_{\pm 2.2}$ & 99$_{\pm 2.2}$ & 96$_{\pm 2.2}$ & 97$_{\pm 4.5}$ & 98$_{\pm 2.7}$ & 98$_{\pm 2.7}$ & 96$_{\pm 4.2}$ & 98$_{\pm 2.7}$ & 85$_{\pm 5.0}$ \\
\rowcolor[HTML]{EFEFEF}
RVT-2~\cite{goyal2024rvt2}  & 94$_{\pm 4.2}$ & \textbf{99}$_{\pm 2.2}$ & 98$_{\pm 2.7}$ & \textbf{100}$_{\pm 0.0}$ & \textbf{99}$_{\pm 2.2}$ & \textbf{99}$_{\pm 2.2}$ & \textbf{99}$_{\pm 2.2}$ & \textbf{100}$_{\pm 0.0}$ & \textbf{100}$_{\pm 0.0}$ & \textbf{100}$_{\pm 0.0}$ & 88$_{\pm 8.4}$ \\
3D-LOTUS (ours) & 93$_{\pm 6.0}$ & \textbf{99}$_{\pm 2.0}$ & \textbf{100}$_{\pm 0.0}$ & 99$_{\pm 2.0}$ & 97$_{\pm 4.0}$ & 96$_{\pm 3.7}$ & 94$_{\pm 4.9}$ & 99$_{\pm 2.0}$ & 99$_{\pm 2.0}$ & \textbf{100}$_{\pm 0.0}$ & \textbf{89}$_{\pm 5.8}$ \\ 
\rowcolor[HTML]{EFEFEF}
3D-LOTUS++ (ours) & 75$_{\pm 4.5}$ & 97$_{\pm 6.0}$ & 94$_{\pm 3.7}$ & 93$_{\pm 5.1}$ & 86$_{\pm 8.0}$ & 88$_{\pm 6.8}$ & 91$_{\pm 4.9}$ & \textbf{100$_{\pm 0.0}$} & \textbf{100$_{\pm 0.0}$} & \textbf{100$_{\pm 0.0}$} & 1$_{\pm 2.0}$ \\
\midrule

\rowcolor[HTML]{CBCEFB}
Method & \begin{tabular}[c]{@{}c@{}}PutIn\\ Cupboard+3\end{tabular} & \begin{tabular}[c]{@{}c@{}}PutMoney\\ InSafe+0\end{tabular} & \begin{tabular}[c]{@{}c@{}}PutMoney\\ InSafe+1\end{tabular} & \begin{tabular}[c]{@{}c@{}}Reach\&\\ Drag+14\end{tabular} & \begin{tabular}[c]{@{}c@{}}Reach\&\\ Drag+18\end{tabular} & \begin{tabular}[c]{@{}c@{}}Slide\\ Block+0\end{tabular} & \begin{tabular}[c]{@{}c@{}}Slide\\ Block+1\end{tabular} & \begin{tabular}[c]{@{}c@{}}Stack\\ Blocks+30\end{tabular} & \begin{tabular}[c]{@{}c@{}}Stack\\ Blocks+36\end{tabular} & \begin{tabular}[c]{@{}c@{}}Stack\\ Blocks+39\end{tabular} &  \\ \midrule
Hiveformer~\cite{guhur2023hiveformer}  & 74$_{\pm 6.5}$ & 85$_{\pm 3.5}$ & 88$_{\pm 2.7}$ & 37$_{\pm 5.7}$ & 32$_{\pm 7.6}$ & 99$_{\pm 2.2}$ & 91$_{\pm 12.4}$ & 6$_{\pm 5.5}$ & 7$_{\pm 4.5}$ & 6$_{\pm 4.2}$ &  \\
\rowcolor[HTML]{EFEFEF}
PolarNet~\cite{chen2023polarnet}  & \textbf{88}$_{\pm 4.5}$ & 93$_{\pm 4.5}$ & 95$_{\pm 5.0}$ & 99$_{\pm 2.2}$ & 99$_{\pm 2.2}$ & \textbf{100}$_{\pm 0.0}$ & 0$_{\pm 0.0}$ & 34$_{\pm 10.8}$ & 30$_{\pm 9.4}$ & 36$_{\pm 12.9}$ &  \\
3D diffuser actor~\cite{ke20243ddifusseractor}  & 82$_{\pm 11.5}$ & \textbf{95}$_{\pm 5.0}$ & 98$_{\pm 2.7}$ & \textbf{100}$_{\pm 0.0}$ & 99$_{\pm 2.2}$ & \textbf{100}$_{\pm 0.0}$ & 89$_{\pm 4.2}$ & 88$_{\pm 7.6}$ & 85$_{\pm 6.1}$ & 89$_{\pm 5.5}$ &  \\
\rowcolor[HTML]{EFEFEF}
RVT-2~\cite{goyal2024rvt2}  & 80$_{\pm 6.1}$ & 93$_{\pm 8.4}$ & 96$_{\pm 8.5}$ & 85$_{\pm 10.0}$ & 94$_{\pm 2.2}$ & \textbf{100}$_{\pm 0.0}$ & 37$_{\pm 6.7}$ & 88$_{\pm 5.7}$ & \textbf{93}$_{\pm 2.7}$ & 88$_{\pm 11.5}$ &  \\
3D-LOTUS (ours) & 72$_{\pm 11.2}$ & 94$_{\pm 3.7}$ & \textbf{99}$_{\pm 2.0}$ & 99$_{\pm 2.0}$ & \textbf{100}$_{\pm 0.0}$ & \textbf{100}$_{\pm 0.0}$ & \textbf{100}$_{\pm 0.0}$ & \textbf{94}$_{\pm 5.8}$ & 91$_{\pm 6.6}$ & \textbf{90}$_{\pm 4.5}$ & \\ 
\rowcolor[HTML]{EFEFEF}
3D-LOTUS++ (ours) & 2$_{\pm 2.5}$ & 22$_{\pm 6.8}$ & 16$_{\pm 4.9}$ & 94$_{\pm 3.7}$ & 62$_{\pm 8.7}$ & \textbf{100$_{\pm 0.0}$} & 65$_{\pm 5.5}$ & 86$_{\pm 5.8}$ & 20$_{\pm 4.5}$ & 28$_{\pm 13.6}$ & \\
\bottomrule

\end{tabular}
\end{table*}

\begin{table*}[t]
\tabcolsep=0.16cm
\caption{Performance on GemBench Level 2.}
\label{tab:gembench_sota_cmpr_l2_detail}
\begin{tabular}{lcccccccccc} \toprule
\rowcolor[HTML]{CBCEFB}
Method & Avg. & \begin{tabular}[c]{@{}c@{}}Push\\ Button+13\end{tabular} & \begin{tabular}[c]{@{}c@{}}Push\\ Button+15\end{tabular} & \begin{tabular}[c]{@{}c@{}}Push\\ Button+17\end{tabular} & \begin{tabular}[c]{@{}c@{}}Pick\&\\ Lift+14\end{tabular} & \begin{tabular}[c]{@{}c@{}}Pick\&\\ Lift+16\end{tabular} & \begin{tabular}[c]{@{}c@{}}Pick\&\\ Lift+18\end{tabular} & \begin{tabular}[c]{@{}c@{}}PickUp\\ Cup+10\end{tabular} & \begin{tabular}[c]{@{}c@{}}PickUp\\ Cup+12\end{tabular} & \begin{tabular}[c]{@{}c@{}}PickUp\\ Cup+13\end{tabular} \\ \midrule
Hiveformer & 26.1$_{\pm 1.4}$ & 97$_{\pm 2.7}$ & 85$_{\pm 10.0}$ & 88$_{\pm 2.7}$ & 21$_{\pm 6.5}$ & 9$_{\pm 4.2}$ & 8$_{\pm 6.7}$ & 30$_{\pm 7.1}$ & 22$_{\pm 13.5}$ & 26$_{\pm 10.6}$ \\
\rowcolor[HTML]{EFEFEF}
PolarNet & 37.1$_{\pm 1.4}$ & 100$_{\pm 0.0}$ & \textbf{100$_{\pm 0.0}$} & 85$_{\pm 7.9}$ & 3$_{\pm 4.5}$ & 1$_{\pm 2.2}$ & 0$_{\pm 0.0}$ & 48$_{\pm 11.0}$ & 46$_{\pm 8.9}$ & 16$_{\pm 6.5}$ \\
3D diffuser actor & 43.4$_{\pm 2.8}$ & 87$_{\pm 13.0}$ & 81$_{\pm 6.5}$ & 60$_{\pm 9.4}$ & 9$_{\pm 4.2}$ & 18$_{\pm 9.1}$ & 0$_{\pm 0.0}$ & 84$_{\pm 5.5}$ & 60$_{\pm 11.7}$ & 62$_{\pm 13.0}$ \\
\rowcolor[HTML]{EFEFEF}
RVT-2 & 51.0$_{\pm 2.3}$ & \textbf{100$_{\pm 0.0}$} & \textbf{100$_{\pm 0.0}$} & \textbf{100$_{\pm 0.0}$} & 47$_{\pm 7.6}$ & 29$_{\pm 9.6}$ & 8$_{\pm 4.5}$ & 81$_{\pm 8.2}$ & 59$_{\pm 9.6}$ & 72$_{\pm 9.7}$ \\
3D-LOTUS (ours) & 49.9$_{\pm 2.2}$ & 99$_{\pm 2.0}$ & \textbf{100$_{\pm 0.0}$} & \textbf{100$_{\pm 0.0}$} & 3$_{\pm 2.5}$ & 18$_{\pm 8.7}$ & 33$_{\pm 9.3}$ & \textbf{89$_{\pm 3.7}$} & 78$_{\pm 8.7}$ & 57$_{\pm 7.5}$ \\ 
\rowcolor[HTML]{EFEFEF}
3D-LOTUS++ (ours) & \textbf{64.5$_{\pm 0.9}$} & 99$_{\pm 2.0}$ & \textbf{100$_{\pm 0.0}$} & 99$_{\pm 2.0}$ & \textbf{94$_{\pm 3.7}$} & \textbf{96$_{\pm 3.7}$} & \textbf{95$_{\pm 3.2}$} & 79$_{\pm 4.9}$ & \textbf{89$_{\pm 9.7}$} & \textbf{84$_{\pm 10.2}$} \\ 
\midrule

\rowcolor[HTML]{CBCEFB}
Method & \begin{tabular}[c]{@{}c@{}}Stack\\ Blocks+24\end{tabular} & \begin{tabular}[c]{@{}c@{}}Stack\\ Blocks+27\end{tabular} & \begin{tabular}[c]{@{}c@{}}Stack\\ Blocks+33\end{tabular} & \begin{tabular}[c]{@{}c@{}}Slide\\ Block+2\end{tabular} & \begin{tabular}[c]{@{}c@{}}Slide\\ Block+3\end{tabular} & \begin{tabular}[c]{@{}c@{}}Close\\ Jar+3\end{tabular} & \begin{tabular}[c]{@{}c@{}}Close\\ Jar+4\end{tabular} & \begin{tabular}[c]{@{}c@{}}LightBulb\\ In+1\end{tabular} & \begin{tabular}[c]{@{}c@{}}LightBulb\\ In+2\end{tabular} & \begin{tabular}[c]{@{}c@{}}Lamp\\ On+0\end{tabular} \\ \midrule
Hiveformer & 0$_{\pm 0.0}$ & 4$_{\pm 4.2}$ & 0$_{\pm 0.0}$ & 0$_{\pm 0.0}$ & 0$_{\pm 0.0}$ & 0$_{\pm 0.0}$ & 0$_{\pm 0.0}$ & 4$_{\pm 4.2}$ & 0$_{\pm 0.0}$ & 7$_{\pm 4.5}$ \\
\rowcolor[HTML]{EFEFEF}
PolarNet & 1$_{\pm 2.2}$ & 2$_{\pm 2.7}$ & 6$_{\pm 8.2}$ & 0$_{\pm 0.0}$ & 0$_{\pm 0.0}$ & 20$_{\pm 10.6}$ & 82$_{\pm 5.7}$ & 22$_{\pm 11.5}$ & 17$_{\pm 8.4}$ & \textbf{14}$_{\pm 10.8}$ \\
3D diffuser actor & \textbf{66}$_{\pm 13.9}$ & 82$_{\pm 2.7}$ & 50$_{\pm 14.6}$ & 0$_{\pm 0.0}$ & 0$_{\pm 0.0}$ & 23$_{\pm 16.8}$ & 82$_{\pm 5.7}$ & 51$_{\pm 17.8}$ & \textbf{60}$_{\pm 10.0}$ & 7$_{\pm 7.6}$ \\
\rowcolor[HTML]{EFEFEF}
RVT-2 & 18$_{\pm 4.5}$ & 56$_{\pm 16.7}$ & 45$_{\pm 13.7}$ & 0$_{\pm 0.0}$ & 1$_{\pm 2.2}$ & 7$_{\pm 7.6}$ & 77$_{\pm 5.7}$ & \textbf{68}$_{\pm 14.4}$ & 6$_{\pm 6.5}$ & 0$_{\pm 0.0}$ \\
3D-LOTUS (ours) & 13$_{\pm 8.1}$ & 40$_{\pm 9.5}$ & \textbf{69}$_{\pm 5.8}$ & 0$_{\pm 0.0}$ & 0$_{\pm 0.0}$ & 71$_{\pm 5.8}$ & 90$_{\pm 4.5}$ & 24$_{\pm 4.9}$ & 41$_{\pm 8.6}$ & 0$_{\pm 0.0}$ \\ 
\rowcolor[HTML]{EFEFEF}
3D-LOTUS++ (ours) & 22$_{\pm 9.3}$ & \textbf{83}$_{\pm 7.5}$ & 59$_{\pm 3.7}$ & \textbf{27}$_{\pm 9.8}$ & \textbf{5}$_{\pm 3.2}$ & \textbf{98}$_{\pm 2.5}$ & \textbf{96}$_{\pm 3.7}$ & 56$_{\pm 9.7}$ & 43$_{\pm 7.5}$ & 2$_{\pm 2.0}$ \\ 
\midrule

\rowcolor[HTML]{CBCEFB}
Method & \begin{tabular}[c]{@{}c@{}}Reach\&\\ Drag+5\end{tabular} & \begin{tabular}[c]{@{}c@{}}Reach\&\\ Drag+7\end{tabular} & \begin{tabular}[c]{@{}c@{}}PutCube\\ InSafe+0\end{tabular} & \begin{tabular}[c]{@{}c@{}}Pick\&Lift\\ Cylinder+0\end{tabular} & \begin{tabular}[c]{@{}c@{}}Pick\&Lift\\ Star+0\end{tabular} & \begin{tabular}[c]{@{}c@{}}Pick\&Lift\\ Moon+0\end{tabular} & \begin{tabular}[c]{@{}c@{}}Pick\&Lift\\ Toy+0\end{tabular} & \begin{tabular}[c]{@{}c@{}}PutIn\\ Cupboard+7\end{tabular} & \begin{tabular}[c]{@{}c@{}}PutIn\\ Cupboard+8\end{tabular} &  \\ \midrule
Hiveformer & 1$_{\pm 2.2}$ & 0$_{\pm 0.0}$ & 4$_{\pm 2.2}$ & 78$_{\pm 5.7}$ & 73$_{\pm 7.6}$ & 88$_{\pm 2.7}$ & 87$_{\pm 4.5}$ & 0$_{\pm 0.0}$ & 0$_{\pm 0.0}$ &  \\
\rowcolor[HTML]{EFEFEF}
PolarNet & 61$_{\pm 8.2}$ & 10$_{\pm 6.1}$ & \textbf{40}$_{\pm 14.1}$ & 93$_{\pm 6.7}$ & 88$_{\pm 8.4}$ & 93$_{\pm 6.7}$ & 90$_{\pm 3.5}$ & 0$_{\pm 0.0}$ & 0$_{\pm 0.0}$ &  \\
3D diffuser actor & 0$_{\pm 0.0}$ & 64$_{\pm 6.5}$ & 3$_{\pm 2.7}$ & \textbf{99}$_{\pm 2.2}$ & 43$_{\pm 17.9}$ & 91$_{\pm 9.6}$ & 30$_{\pm 9.4}$ & 0$_{\pm 0.0}$ & \textbf{3}$_{\pm 4.5}$ &  \\
\rowcolor[HTML]{EFEFEF}
RVT-2 & 91$_{\pm 2.2}$ & \textbf{89}$_{\pm 6.5}$ & 6$_{\pm 5.5}$ & 98$_{\pm 2.7}$ & \textbf{98}$_{\pm 4.5}$ & \textbf{94}$_{\pm 4.2}$ & 78$_{\pm 8.4}$ & 0$_{\pm 0.0}$ & 0$_{\pm 0.0}$ &  \\
3D-LOTUS (ours) & \textbf{95}$_{\pm 4.5}$ & 18$_{\pm 10.8}$ & 25$_{\pm 5.5}$ & 88$_{\pm 8.7}$ & 69$_{\pm 6.6}$ & 80$_{\pm 8.4}$ & \textbf{96}$_{\pm 3.7}$ & 0$_{\pm 0.0}$ & 0$_{\pm 0.0}$  & \\ 
\rowcolor[HTML]{EFEFEF}
3D-LOTUS++ (ours) & 94$_{\pm 2.0}$ & 64$_{\pm 12.4}$ & 37$_{\pm 5.1}$ & 91$_{\pm 2.0}$ & 94$_{\pm 3.7}$ & 29$_{\pm 6.6}$ & 71$_{\pm 2.0}$ & \textbf{1}$_{\pm 2.0}$ & 0$_{\pm 0.0}$  & \\ \bottomrule
\end{tabular}
\end{table*}

\begin{table*}
\centering
\tabcolsep=0.22cm
\caption{Performance on GemBench Level 3.}
\label{tab:gembench_sota_cmpr_l3_detail}
\begin{tabular}{lcccccccc} \toprule
\rowcolor[HTML]{CBCEFB}
Method & Avg. & \begin{tabular}[c]{@{}c@{}}Close\\ Door+0\end{tabular} & \begin{tabular}[c]{@{}c@{}}Close\\ Box+0\end{tabular} & \begin{tabular}[c]{@{}c@{}}Close\\ Fridge2+0\end{tabular} & \begin{tabular}[c]{@{}c@{}}CloseLaptop\\ Lid2+0\end{tabular} & \begin{tabular}[c]{@{}c@{}}Close\\ Microwave2+0\end{tabular} & \begin{tabular}[c]{@{}c@{}}Open\\ Door2+0\end{tabular} & \begin{tabular}[c]{@{}c@{}}Open\\ Box2+0\end{tabular} \\
Hiveformer & 35.1$_{\pm 1.7}$ & 0$_{\pm 0.0}$ & 1$_{\pm 2.2}$ & 34$_{\pm 9.6}$ & 52$_{\pm 9.1}$ & 15$_{\pm 7.1}$ & 32$_{\pm 11.5}$ & 5$_{\pm 3.5}$ \\
\rowcolor[HTML]{EFEFEF}
PolarNet & 38.5$_{\pm 1.7}$ & 0$_{\pm 0.0}$ & 0$_{\pm 0.0}$ & 78$_{\pm 5.7}$ & 26$_{\pm 8.2}$ & 74$_{\pm 6.5}$ & 33$_{\pm 6.7}$ & 23$_{\pm 8.4}$ \\
3D diffuser actor & 37.0$_{\pm 2.2}$ & 0$_{\pm 0.0}$ & 0$_{\pm 0.0}$ & \textbf{97}$_{\pm 2.7}$ & 23$_{\pm 6.7}$ & 88$_{\pm 7.6}$ & \textbf{86}$_{\pm 7.4}$ & \textbf{67}$_{\pm 9.8}$ \\
\rowcolor[HTML]{EFEFEF}
RVT-2 & 36.0$_{\pm 2.2}$ & 1$_{\pm 2.2}$ & 2$_{\pm 2.7}$ & 72$_{\pm 6.7}$ & 42$_{\pm 14.0}$ & 71$_{\pm 8.9}$ & 79$_{\pm 6.5}$ & 5$_{\pm 6.1}$ \\
3D-LOTUS (ours) & 38.1$_{\pm 1.1}$ & 0$_{\pm 0.0}$ & \textbf{58}$_{\pm 8.1}$ & 36$_{\pm 9.7}$ & \textbf{54}$_{\pm 10.7}$ & 85$_{\pm 7.1}$ & 42$_{\pm 6.8}$ & 11$_{\pm 6.6}$ \\
\rowcolor[HTML]{EFEFEF}
3D-LOTUS++ (ours) & \textbf{41.5}$_{\pm 1.8}$ & \textbf{1}$_{\pm 2.0}$ & 29$_{\pm 8.6}$ & 93$_{\pm 2.5}$ & 50$_{\pm 9.5}$ & \textbf{99}$_{\pm 2.0}$ & 52$_{\pm 10.3}$ & 16$_{\pm 8.0}$ \\
\midrule
\rowcolor[HTML]{CBCEFB}
Method & \begin{tabular}[c]{@{}c@{}}Open\\ Drawer2+0\end{tabular} & \begin{tabular}[c]{@{}c@{}}Open\\ Drawer3+0\end{tabular} & \begin{tabular}[c]{@{}c@{}}OpenDrawer\\ Long+0\end{tabular} & \begin{tabular}[c]{@{}c@{}}OpenDrawer\\ Long+1\end{tabular} & \begin{tabular}[c]{@{}c@{}}OpenDrawer\\ Long+2\end{tabular} & \begin{tabular}[c]{@{}c@{}}OpenDrawer\\ Long+3\end{tabular} & \begin{tabular}[c]{@{}c@{}}Toilet\\ SeatUp+0\end{tabular} & \begin{tabular}[c]{@{}c@{}}Open\\ Fridge+0\end{tabular} \\
Hiveformer & 59$_{\pm 11.9}$ & 39$_{\pm 11.9}$ & 78$_{\pm 8.4}$ & 82$_{\pm 4.5}$ & 49$_{\pm 4.2}$ & 57$_{\pm 11.5}$ & 6$_{\pm 4.2}$ & 0$_{\pm 0.0}$ \\
\rowcolor[HTML]{EFEFEF}
PolarNet & \textbf{91}$_{\pm 4.2}$ & 29$_{\pm 8.2}$ & 84$_{\pm 11.9}$ & \textbf{88}$_{\pm 5.7}$ & \textbf{63}$_{\pm 8.4}$ & 37$_{\pm 7.6}$ & 2$_{\pm 2.7}$ & 4$_{\pm 2.2}$ \\
3D diffuser actor & 19$_{\pm 8.2}$ & 1$_{\pm 2.2}$ & 15$_{\pm 5.0}$ & 35$_{\pm 13.7}$ & 26$_{\pm 9.6}$ & \textbf{79}$_{\pm 12.9}$ & 0$_{\pm 0.0}$ & \textbf{7}$_{\pm 5.7}$ \\
\rowcolor[HTML]{EFEFEF}
RVT-2 & 81$_{\pm 11.9}$ & 0$_{\pm 0.0}$ & \textbf{84}$_{\pm 8.2}$ & 39$_{\pm 10.8}$ & 11$_{\pm 8.9}$ & 75$_{\pm 6.1}$ & 7$_{\pm 5.7}$ & 0$_{\pm 0.0}$ \\
3D-LOTUS (ours) & 90$_{\pm 3.2}$ & 22$_{\pm 8.1}$ & 56$_{\pm 13.9}$ & 33$_{\pm 11.2}$ & 17$_{\pm 8.1}$ & 75$_{\pm 6.3}$ & 0$_{\pm 0.0}$ & 4$_{\pm 5.8}$ \\
\rowcolor[HTML]{EFEFEF}
3D-LOTUS++ (ours) & 70$_{\pm 5.5}$ & \textbf{41}$_{\pm 4.9}$ & 72$_{\pm 4.0}$ & 52$_{\pm 10.8}$ & 23$_{\pm 8.1}$ & 78$_{\pm 5.1}$ & \textbf{8}$_{\pm 5.1}$ & 0$_{\pm 0.0}$ \\
\midrule

\rowcolor[HTML]{CBCEFB}
Method & \begin{tabular}[c]{@{}c@{}}OpenLaptop\\ Lid+0\end{tabular} & \begin{tabular}[c]{@{}c@{}}Open\\ Microwave+0\end{tabular} & \begin{tabular}[c]{@{}c@{}}PutMoney\\ InSafe+2\end{tabular} & \begin{tabular}[c]{@{}c@{}}Open\\ Drawer+1\end{tabular} & \begin{tabular}[c]{@{}c@{}}Close\\ Drawer+0\end{tabular} & \begin{tabular}[c]{@{}c@{}}Close\\ Grill+0\end{tabular} &  &  \\
Hiveformer & \textbf{100}$_{\pm 0.0}$ & 0$_{\pm 0.0}$ & 0$_{\pm 0.0}$ & 0$_{\pm 0.0}$ & 83$_{\pm 5.7}$ & 44$_{\pm 10.8}$ &  &  \\
\rowcolor[HTML]{EFEFEF}
PolarNet & \textbf{100}$_{\pm 0.0}$ & 0$_{\pm 0.0}$ & 1$_{\pm 2.2}$ & 4$_{\pm 4.2}$ & 29$_{\pm 11.9}$ & 42$_{\pm 11.5}$ &  &  \\
3D diffuser actor & \textbf{100}$_{\pm 0.0}$ & 0$_{\pm 0.0}$ & 2$_{\pm 4.5}$ & 0$_{\pm 0.0}$ & 66$_{\pm 7.4}$ & \textbf{65}$_{\pm 13.7}$ &  &  \\
\rowcolor[HTML]{EFEFEF}
RVT-2 & 93$_{\pm 5.7}$ & 0$_{\pm 0.0}$ & 0$_{\pm 0.0}$ & \textbf{6}$_{\pm 2.2}$ & 78$_{\pm 8.4}$ & 9$_{\pm 4.2}$ &  &  \\
3D-LOTUS (ours) & \textbf{100}$_{\pm 0.0}$ & 0$_{\pm 0.0}$ & 0$_{\pm 0.0}$ & 0$_{\pm 0.0}$ & \textbf{87}$_{\pm 8.1}$ & 29$_{\pm 6.6}$ &  &  \\
\rowcolor[HTML]{EFEFEF}
3D-LOTUS++ (ours) & 86$_{\pm 6.6}$ & 0$_{\pm 0.0}$ & \textbf{13}$_{\pm 8.1}$ & 0$_{\pm 0.0}$ & 69$_{\pm 5.8}$ & 19$_{\pm 13.9}$ &  &  \\
\bottomrule
\end{tabular}
\end{table*}

\begin{table*}[t]
\centering
\tabcolsep=0.2cm
\caption{Performance on GemBench Level 4.}
\label{tab:gembench_sota_cmpr_l4_detail}
\begin{tabular}{lccccccc} \toprule
\rowcolor[HTML]{CBCEFB}
Method & Avg. & \begin{tabular}[c]{@{}c@{}}Push\\ Buttons4+1\end{tabular} & \begin{tabular}[c]{@{}c@{}}Push\\ Buttons4+2\end{tabular} & \begin{tabular}[c]{@{}c@{}}Push\\ Buttons4+3\end{tabular} & \begin{tabular}[c]{@{}c@{}}TakeShoes\\ OutOfBox+0\end{tabular} & \begin{tabular}[c]{@{}c@{}}PutItems\\ InDrawer+0\end{tabular} & \begin{tabular}[c]{@{}c@{}}PutItems\\ InDrawer+2\end{tabular} \\
Hiveformer & 0$_{\pm 0.0}$ & 0$_{\pm 0.0}$ & 0$_{\pm 0.0}$ & 0$_{\pm 0.0}$ & 0$_{\pm 0.0}$ & 0$_{\pm 0.0}$ & 0$_{\pm 0.0}$ \\
\rowcolor[HTML]{EFEFEF}
PolarNet & 0.1$_{\pm 0.2}$ & 1$_{\pm 2.2}$ & 0$_{\pm 0.0}$ & 0$_{\pm 0.0}$ & 0$_{\pm 0.0}$ & 0$_{\pm 0.0}$ & 0$_{\pm 0.0}$ \\
3D diffuser actor & 0$_{\pm 0.0}$ & 0$_{\pm 0.0}$ & 0$_{\pm 0.0}$ & 0$_{\pm 0.0}$ & 0$_{\pm 0.0}$ & 0$_{\pm 0.0}$ & 0$_{\pm 0.0}$ \\
\rowcolor[HTML]{EFEFEF}
RVT-2 & 0$_{\pm 0.0}$ & 0$_{\pm 0.0}$ & 0$_{\pm 0.0}$ & 0$_{\pm 0.0}$ & 0$_{\pm 0.0}$ & 0$_{\pm 0.0}$ & 0$_{\pm 0.0}$ \\
3D-LOTUS (ours) & 0.3$_{\pm 0.3}$ & 3$_{\pm 4.0}$ & 0$_{\pm 0.0}$ & 0$_{\pm 0.0}$ & 0$_{\pm 0.0}$ & 0$_{\pm 0.0}$ & 0$_{\pm 0.0}$ \\
\rowcolor[HTML]{EFEFEF}
3D-LOTUS++ (ours) & \textbf{17.4}$_{\pm 0.4}$ & \textbf{76}$_{\pm 7.4}$ & \textbf{49}$_{\pm 8.6}$ & \textbf{37}$_{\pm 8.1}$ & 0$_{\pm 0.0}$ & 0$_{\pm 0.0}$ & 0$_{\pm 0.0}$ \\
\midrule
\rowcolor[HTML]{CBCEFB}
Method & \begin{tabular}[c]{@{}c@{}}PutItems\\ InDrawer+4\end{tabular} & Tower4+1 & Tower4+3 & \begin{tabular}[c]{@{}c@{}}Stack\\ Cups+0\end{tabular} & \begin{tabular}[c]{@{}c@{}}Stack\\ Cups+3\end{tabular} & \begin{tabular}[c]{@{}c@{}}PutAllGroceries\\ InCupboard+0\end{tabular} &  \\
Hiveformer & 0$_{\pm 0.0}$ & 0$_{\pm 0.0}$ & 0$_{\pm 0.0}$ & 0$_{\pm 0.0}$ & 0$_{\pm 0.0}$ & 0$_{\pm 0.0}$ &  \\
\rowcolor[HTML]{EFEFEF}
PolarNet & 0$_{\pm 0.0}$ & 0$_{\pm 0.0}$ & 0$_{\pm 0.0}$ & 0$_{\pm 0.0}$ & 0$_{\pm 0.0}$ & 0$_{\pm 0.0}$ &  \\
3D diffuser actor & 0$_{\pm 0.0}$ & 0$_{\pm 0.0}$ & 0$_{\pm 0.0}$ & 0$_{\pm 0.0}$ & 0$_{\pm 0.0}$ & 0$_{\pm 0.0}$ &  \\
\rowcolor[HTML]{EFEFEF}
RVT-2 & 0$_{\pm 0.0}$ & 0$_{\pm 0.0}$ & 0$_{\pm 0.0}$ & 0$_{\pm 0.0}$ & 0$_{\pm 0.0}$ & 0$_{\pm 0.0}$ &  \\
3D-LOTUS (ours) & 0$_{\pm 0.0}$ & 0$_{\pm 0.0}$ & 0$_{\pm 0.0}$ & 0$_{\pm 0.0}$ & 0$_{\pm 0.0}$ & 0$_{\pm 0.0}$ &  \\
\rowcolor[HTML]{EFEFEF}
3D-LOTUS++ (ours) & 0$_{\pm 0.0}$ & \textbf{17}$_{\pm 10.8}$ & \textbf{30}$_{\pm 13.4}$ & 0$_{\pm 0.0}$ & 0$_{\pm 0.0}$ & 0$_{\pm 0.0}$ & \\
\bottomrule
\end{tabular}
\end{table*}

\subsection{Detail task specification}
We describe each task in detail along with its variations below, and highlight the newly created tasks in GemBench.

\noindent\textbf{Push Button}

\textbf{Filename:} push\_button.py

\textbf{Task:} Push button with the specified color.

\textbf{New/Modified:} No.

\textbf{Variations per level:} Level 1: 0 (maroon), 3 (navy) and 4 (yellow). Level 2: 13 (azure), 15 (rose) and 17 (white).

\textbf{Objects:} 1 button.

\textbf{Success Metric:} The button is completely pressed.
\newline

\noindent\textbf{Close Fridge}

\textbf{Filename:} close\_fridge.py

\textbf{Task:} Close the fridge door.

\textbf{New/Modified:} No.

\textbf{Objects:} 1 fridge.

\textbf{Variations per level:} Level 1: 0.

\textbf{Success Metric:} The revolute joint of the fridge door has been rotated so that the door is closed and in contact with the fridge cabinet.
\newline

\noindent\textbf{Close Laptop Lid}

\textbf{Filename:} close\_laptop\_lid.py

\textbf{Task:} Grasp the laptop lid and rotate to close the laptop.

\textbf{New/Modified:} No.

\textbf{Objects:} 1 box, 1 laptop.

\textbf{Variations per level:} Level 1: 0.

\textbf{Success Metric:} The revolute joint of the laptop lid has been rotated so that the laptop is closed.
\newline

\noindent\textbf{Close Microwave}

\textbf{Filename:} close\_microwave.py

\textbf{Task:} Close the microwave door.

\textbf{New/Modified:} No.

\textbf{Objects:} 1 microwave.

\textbf{Variations per level:} Level 1: 0.

\textbf{Success Metric:}  The revolute joint of the microwave has been rotated so that the microwave door is closed.
\newline

\noindent\textbf{Open Door}

\textbf{Filename:} open\_door.py

\textbf{Task:} Pick up the door handle and open the door by pushing.

\textbf{New/Modified:} Yes, modified to add new instructions.

\textbf{Objects:} 1 door with a handle.

\textbf{Variations per level:} Level 1: 0.

\textbf{Success Metric:}  The handle has been rotated 25º to unlock the door and the revolute joint of the door has been rotated 25º.
\newline

\noindent\textbf{Open Box}

\textbf{Filename:} open\_box.py

\textbf{Task:} Open the box lid.

\textbf{New/Modified:} Yes, modified to add new instructions.

\textbf{Objects:} 1 box.

\textbf{Variations per level:} Level 1: 0.

\textbf{Success Metric:}  The revolute joint of the box lid has rotated 90º so that it is in the open configuration.
\newline

\noindent\textbf{Open Drawer}

\textbf{Filename:} open\_drawer.py

\textbf{Task:} Open one of the three drawers: top, middle, or bottom.

\textbf{New/Modified:} No.

\textbf{Objects:} 1 drawer.

\textbf{Variations per level:} Level 1: 0 (bottom) and 2 (top) and Level 3: 1(middle).

\textbf{Success Metric:} The prismatic joint of the specified drawer is fully extended.
\newline

\noindent\textbf{Pick and Lift}

\textbf{Filename:} pick\_and\_lift.py

\textbf{Task:} Pick a colored cube and lift it to a red sphere target.

\textbf{New/Modified:} No.

\textbf{Objects:} 3 colored cubes, one with the specified color and the other two with different colors as distractors.

\textbf{Variations per level:} Level 1: 0 (red), 2 (lime) and 7 (cyan) and Level 2: 14 (teal), 16 (violet) and 18 (black).

\textbf{Success Metric:}  The cube of the specified color is grasped and next to the target red sphere.
\newline

\noindent\textbf{Pick Up Cup}

\textbf{Filename:} pick\_up\_cup.py

\textbf{Task:} Pick the cup with the specified color and lift it from the table.

\textbf{New/Modified:} No.

\textbf{Objects:} 3 tall colored cups.

\textbf{Variations per level:} Level 1: 8 (magenta), 9 (silver) and 11 (orange) and Level 2: 10 (gray), 12 (olive) and 13 (purple).

\textbf{Success Metric:}  The cup of the specified color is grasped and lifted from the table.
\newline

\noindent\textbf{Stack Blocks}

\textbf{Filename:} stack\_blocks.py

\textbf{Task:} Stack $N$ blocks of the specified color on the green platform. There are always 4 blocks of the specified color, and 4 distractor blocks of another color. The block colors are sampled from the full set of 20 color instances.

\textbf{New/Modified:} No.

\textbf{Objects:} 8 color blocks (4 are distractors), and 1 green platform

\textbf{Variations per level:} Level 1: 30 (2 gray blocks), 36 (2 olive blocks), 39 (2 purple blocks) and Level 2: 24 (2 orange blocks), 27 (2 silver blocks) and 33 (2 magenta blocks).

\textbf{Success Metric:} $N$ blocks are inside the area of the green platform.
\newline

\noindent\textbf{Put Groceries in Cupboard}

\textbf{Filename:} put\_groceries\_in\_cupboard.py

\textbf{Task:} Grab the specified object and put it in the cupboard above. The scene always contains 9 YCB objects that are randomly placed on the tabletop.

\textbf{New/Modified:} Yes, modified the object names to include object category and therefore the instructions.

\textbf{Objects:} 9 YCB objects, and 1 cupboard (that hovers in the air like magic).

\textbf{Variations per level:} Level 1: 0 (crackers box) and 3 (soup can) and Level 2: 7 (mustard bottle) and 8 (sugar box).

\textbf{Success Metric:}  The specified object is inside the cupboard.
\newline

\noindent\textbf{Put Money in Safe}

\textbf{Filename:} put\_money\_in\_safe.py

\textbf{Task:} Pick up the stack of money and put it inside the safe on the specified shelf. The shelf has three placement locations: top, middle, bottom.

\textbf{New/Modified:} No.

\textbf{Objects:} 1 stack of money, and 1 safe.

\textbf{Variations per level:} Level 1: 0 (bottom) and 1 (middle) and Level 2: 2 (top).

\textbf{Success Metric:}  The stack of money is on the specified shelf inside the safe.
\newline

\noindent\textbf{Slide Block to Color Target}

\textbf{Filename:} slide\_block\_to\_color\_target\_peract.py

\textbf{Task:} Slide the block on to one of the colored square targets. The target colors are limited to red, blue, pink, and yellow.

\textbf{New/Modified:} Yes, modified as in RLBench-18Task~\cite{shridhar2023peract}. The original slide block to target.py task contained only one target. Three other targets were added to make a total of 4 variations.

\textbf{Objects:} 1 block and 4 colored target squares.

\textbf{Variations per level:} Level 1: 0 (green) and 1 (blue) and Level 2: 2 (pink) and 3 (yellow).

\textbf{Success Metric:}  Some part of the block is inside the specified target area.
\newline

\noindent\textbf{Reach and Drag}

\textbf{Filename:} reach\_and\_drag\_peract.py

\textbf{Task:} Grab the stick and use it to drag the cube on to the specified colored target square. The target colors are sampled from the full set of 20 color instances.

\textbf{New/Modified:} Yes, modified as in RLBench-18Task~\cite{shridhar2023peract}. The original reach\_and\_drag.py task contained only one target. Three other targets were added with randomized colors.

\textbf{Objects:} 1 block, 1 stick, and 4 colored target squares.

\textbf{Variations per level:} Level 1: 14 (teal) and 18 (black) and Level 2: 5 (cyan) and 7 (navy).

\textbf{Success Metric:} Some part of the block is inside the specified target area.
\newline

\noindent\textbf{Close Jar}

\textbf{Filename:} close\_jar.py

\textbf{Task:} Put the lid on the jar with the specified color and screw the lid in. The jar colors are sampled from the full set of 20 color instances.

\textbf{New/Modified:} No.

\textbf{Objects:} 1 jar lid and 2 colored jars.

\textbf{Variations per level:} Level 1: 15 (azure) and 16 (violet) and Level 2: 3 (blue) and 4 (green).

\textbf{Success Metric:}  The lid is on top of the specified jar and the Franka gripper is not grasping anything.
\newline

\noindent\textbf{Screw Bulb}

\textbf{Filename:} light\_bulb\_in.py

\textbf{Task:}  Pick up the light bulb from the specified holder, and screw it into the lamp stand. The colors of holder are sampled from the full set of 20 color instances. There are always two holders in the scene – one specified and one distractor holder.

\textbf{Modified:} No.

\textbf{Objects:} 2 light bulbs, 2 holders, and 1 lamp stand.

\textbf{Variations per level:} Level 1: 17 (rose) and 19 (white) and Level 2: 1 (lime) and 2 (maroon).

\textbf{Success Metric:}  The bulb from the specified holder is inside the lamp stand dock.
\newline

\noindent\textbf{Close Door}

\textbf{Filename:} close\_door.py

\textbf{Task:} Grab the handle of the door and pull to close the door.

\textbf{Modified:} Yes, modified to add new instructions.

\textbf{Objects:} 1 door with a handle.

\textbf{Variations per level:} Level 3: 0.

\textbf{Success Metric:}  The revolute joint of the door is in the close configuration.
\newline 

\noindent\textbf{Close Box}

\textbf{Filename:} close\_box.py

\textbf{Task:} Grab the box lid and rotate to close the box.

\textbf{New/Modified:} Yes, modified to add new instructions.

\textbf{Objects:} 1 box.

\textbf{Variations per level:} Level 3: 0.

\textbf{Success Metric:}  The revolute joint of the box lid has rotated so that it is in the closed configuration.
\newline

\noindent\textbf{Close Drawer}

\textbf{Filename:} close\_drawer.py

\textbf{Task:} Close one of the three drawers: top, middle, or bottom.

\textbf{New/Modified:} Yes, modified to add new instructions.

\textbf{Objects:} 1 drawer unit with 3 drawers.

\textbf{Variations per level:} Level 3: 0.

\textbf{Success Metric:}  The prismatic joint of the specified drawer is fully contracted.
\newline

\noindent\textbf{Open Fridge}

\textbf{Filename:} open\_fridge.py

\textbf{Task:} Grab the handle of the fridge door and  pull to open the fridge.

\textbf{New/Modified:} Yes, modified to add new instructions.

\textbf{Objects:} 1 fridge.

\textbf{Variations per level:} Level 3: 0.

\textbf{Success Metric:}   The revolute joint of the fridge door has rotated 70 degrees so that the fridge is open.
\newline

\noindent\textbf{Open Laptop Lid}

\textbf{Filename:} open\_laptop\_lid.py

\textbf{Task:} Grab the laptop lid and open it.

\textbf{New/Modified:} Yes, we modified the close laptop lid task to add a new task where the laptop is already close and the goal is to open the lid instead of closing it.

\textbf{Objects:} 1 laptop and 1 box.

\textbf{Variations per level:} Level 3: 0.

\textbf{Success Metric:}  The revolute joint of the laptop lid reaches its minimum value so that the laptop is open.
\newline

\noindent\textbf{Open Microwave}

\textbf{Filename:} open\_microwave.py

\textbf{Task:} Grab the handle to pull and open the microwave door.

\textbf{New/Modified:} Yes, modified to add new instructions.

\textbf{Objects:} 1 microwave.

\textbf{Variations per level:} Level 3: 0.

\textbf{Success Metric:}  The revolute joint of the microwave has rotated at least 80º.
\newline

\noindent\textbf{Put Cube In Safe}

\textbf{Filename:} put\_cube\_in\_safe.py

\textbf{Task:} Pick up the cube and put it inside the safe on the specified shelf. The shelf has three placement locations: top, middle, bottom.

\textbf{New/Modified:} Yes, new task based on put money in safe where the bank note is changed by a cube.

\textbf{Objects:} 1 cube and 1 safe.

\textbf{Variations per level:} Level 1: 0.

\textbf{Success Metric:}  The cube is on the specified shelf inside the safe.
\newline

\noindent\textbf{Close Fridge2}

\textbf{Filename:} close\_fridge2.py

\textbf{Task:} Close the fridge door by pushing it.

\textbf{New/Modified:} Yes, new task based on close fridge where the fridge mesh and texture are completely changed.

\textbf{Objects:} 1 microwave.

\textbf{Variations per level:} Level 1: 0.

\textbf{Success Metric:} The revolute joint of the fridge door is in closed configuration and the door is in contact with the fridge cabinet.
\newline

\noindent\textbf{Close Laptop Lid2}

\textbf{Filename:} close\_laptop\_lid2.py

\textbf{Task:} Grasp the laptop lid and rotate to close the laptop.

\textbf{New/Modified:} Yes, a new task based on close laptop lid task where the laptop is already close and the goal is to open the lid instead of closing it.

\textbf{Objects:} 1 box, 1 laptop.

\textbf{Variations per level:} Level 3: 0.

\textbf{Success Metric:} The revolute joint of the laptop lid is in the close configuration.
\newline

\noindent\textbf{Close Microwave2}

\textbf{Filename:} close\_microwave2.py

\textbf{Task:} Close the microwave door.

\textbf{New/Modified:} Yes, new task based on close microwave task where the microwave and handle meshes and textures are completely changed.

\textbf{Objects:} 1 microwave.

\textbf{Variations per level:} Level 3: 0.

\textbf{Success Metric:}  The revolute joint of the microwave is in the close configuration.
\newline

\noindent\textbf{Open Door2}

\textbf{Filename:} open\_door2.py

\textbf{Task:} Pick up the door handle and open the door by pushing.

\textbf{New/Modified:} Yes, a new task based on open door task where the door, door frame and handle meshes and textures are completely changed.

\textbf{Objects:} 1 door with a handle.

\textbf{Variations per level:} Level 3: 0.

\textbf{Success Metric:}  The handle has been rotated 25º to unlock the door and the revolute joint of the door has been rotated 20º.
\newline

\noindent\textbf{Open Box2}

\textbf{Filename:} open\_box2.py

\textbf{Task:} Open the box lid.

\textbf{New/Modified:} Yes, a new task based on open box task where the box mesh and texture are changed.

\textbf{Objects:} 1 box with a lid.

\textbf{Variations per level:} Level 3: 0.

\textbf{Success Metric:}  The revolute joint of the box lid has rotated 90º so that it is in the open configuration.
\newline

\noindent\textbf{Open Drawer2}

\textbf{Filename:} open\_drawer2.py

\textbf{Task:} Open one of the three drawers: top, middle, or bottom.

\textbf{New/Modified:} Yes, a new task based on open drawer task where the drawer handles and cabinet meshes, colors and textures are changed.

\textbf{Objects:} 1 drawer.

\textbf{Variations per level:} Level 3: 0 (bottom).

\textbf{Success Metric:}  The prismatic joint of the specified drawer is fully extended.
\newline

\noindent\textbf{Open Drawer3}

\textbf{Filename:} open\_drawer3.py

\textbf{Task:} Open one of the three drawers: top, middle, or bottom.

\textbf{New/Modified:} Yes, a new task based on open drawer task where the drawer handles and cabinet meshes, colors and textures are changed and different to Open Drawer2.

\textbf{Objects:} 1 drawer.

\textbf{Variations per level:} Level 3: 0 (bottom).

\textbf{Success Metric:} The prismatic joint of the specified drawer is fully extended.
\newline

\noindent\textbf{Open Drawer Long}

\textbf{Filename:} open\_drawer\_long.py

\textbf{Task:} Open one of the four drawers: top, top middle, bottom middle or bottom.

\textbf{New/Modified:}  Yes, a new task based on open drawer task where the drawer unit has 4 drawers and and the handles and cabinet meshes, colors and textures are changed.

\textbf{Objects:} 1 drawer.

\textbf{Variations per level:} Level 3: 0 (bottom), 1 (bottom middle), 2 (top middle) and 3 (top).

\textbf{Success Metric:} The prismatic joint of the specified drawer is fully extended.
\newline

\noindent\textbf{Lamp On}

\textbf{Filename:} lamp\_on.py

\textbf{Task:} Press the button to light on the lamp.

\textbf{New/Modified:} No.

\textbf{Objects:} 1 lamp, 1 button.

\textbf{Variations per level:} Level 2: 0.

\textbf{Success Metric:}  The button is pressed and the lamp light is on.
\newline

\noindent\textbf{Close Grill}

\textbf{Filename:} close\_grill.py

\textbf{Task:} Close the grill.

\textbf{New/Modified:} No.

\textbf{Objects:} 1 grill.

\textbf{Variations per level:} Level 3: 0.

\textbf{Success Metric:}  The revolute joint of the grill lid is in the close configuration.
\newline

\noindent\textbf{Toilet Seat Up}

\textbf{Filename:} toilet\_seat\_up.py

\textbf{Task:} Grasp the toilet seat and rotate it to move it up the toilet.

\textbf{New/Modified:} No.

\textbf{Objects:} 1 toilet.

\textbf{Variations per level:} Level 3: 0.

\textbf{Success Metric:}  The revolute joint of the toilet seat is in the open configuration.
\newline

\noindent\textbf{Pick and Lift Cylinder}

\textbf{Filename:} pick\_and\_lift\_cylinder.py

\textbf{Task:} Pick and lift the cylinder with the specified color.

\textbf{New/Modified:} Yes, a new task based on pick and lift task where cubes are changed with cylinders.

\textbf{Objects:}  3 colored cylinders, one with the specified color and the other two with different colors as distractors.

\textbf{Variations per level:} Level 2: 0 (red).

\textbf{Success Metric:}  The cylinder of the specified color is grasped and placed in the target red sphere.
\newline

\noindent\textbf{Pick and Lift Star}

\textbf{Filename:} pick\_and\_lift\_star.py

\textbf{Task:} Pick and lift the star with the specified color.

\textbf{New/Modified:} Yes, a new task based on pick and lift task where cubes are changed with stars.

\textbf{Objects:}  3 colored stars, one with the specified color and the other two with different colors as distractors.

\textbf{Variations per level:} Level 2: 0 (red).

\textbf{Success Metric:}  The star of the specified color is grasped and placed in the target red sphere.
\newline

\noindent\textbf{Pick and Lift Moon}

\textbf{Filename:} pick\_and\_lift\_moon.py

\textbf{Task:}  Pick and lift the moon with the specified color.

\textbf{New/Modified:} Yes, a new task based on pick and lift task where cubes are changed with moons.

\textbf{Objects:} 3 colored moons, one with the specified color and the other two with different colors as distractors.

\textbf{Variations per level:} Level 2: 0 (red).

\textbf{Success Metric:}  The moon of the specified color is grasped and placed in the target red sphere.
\newline

\noindent\textbf{Pick and Lift Toy}

\textbf{Filename:} pick\_and\_lift\_moon.py

\textbf{Task:}  Pick and lift the duck rubber toy with the specified color

\textbf{New/Modified:} Yes, a new task based on pick and lift task where cubes are changed with duck rubber toys.

\textbf{Objects:} 3 colored duck rubber toys, one with the specified color and the other two with different colors as distractors.

\textbf{Variations per level:} Level 2: 0 (red).

\textbf{Success Metric:}  The duck rubber toy of the specified color is grasped and placed in the target red sphere.
\newline

\noindent\textbf{Push Buttons4}

\textbf{Filename:} push\_buttons4.py

\textbf{Task:} Push 2/3/4 buttons in the given color order.

\textbf{New/Modified:} Yes, a new task based on push button where there are 4 buttons instead of one, and the goal is change to follow a sequence of 2/3/4 buttons colors.

\textbf{Objects:} 4 colored buttons.

\textbf{Variations per level:} Level 4: 1 (navy then teal), 2 (green then yellow then rose), 3 (maroon then blue then orange then magenta).

\textbf{Success Metric:}  The buttons are pressed consecutively in the same order as specified by the instructions.
\newline

\noindent\textbf{Take Shoes Out Of Box}

\textbf{Filename:} take\_shoes\_out\_of\_box.py

\textbf{Task:} Open the box and then take both shoes and place them in the table.

\textbf{New/Modified:} No.

\textbf{Objects:} 1 box and 2 shoes.

\textbf{Variations per level:} Level 4: 0.

\textbf{Success Metric:} Both shoes are outside of the box and on top of the table.
\newline

\noindent\textbf{Put Items In Drawer}

\textbf{Filename:} put\_items\_in\_drawer.py

\textbf{Task:} Put a cube, a cylinder and a moon in the specified order and inside one of the three drawers: top, middle and bottom.

\textbf{New/Modified:} Yes. The task is a new long horizon task that combines open drawer task and put in drawer.

\textbf{Objects:} 1 cube, 1 cylinder, 1 moon and a drawer.

\textbf{Variations per level:} Level 4: 0 (cube then cylinder then moon in bottom drawer), 2 (cube then cylinder then moon in top drawer) and 4 (cube then moon then cylinder in middle drawer).

\textbf{Success Metric:}  The cube, the cylinder and the moon have been placed in order inside the specified drawer.
\newline

\noindent\textbf{Tower4}

\textbf{Filename:} tower4.py

\textbf{Task:} Stack 2, 3 or 4 cubes to create a tower on top of the green platform.

\textbf{New/Modified:} Yes, a new task based on stack blocks reducing the number of cubes to 4 blocks of different colors without repetitions.

\textbf{Objects:} 4 cubes of different colors and a green platform.

\textbf{Variations per level:} Level 4: 1 (stack white then teal then blue) and 3 (stack orange then gray then lime then rose).

\textbf{Success Metric:}  The cubes are stacked in the correct order and on top of the green platform.
\newline

\noindent\textbf{Stack Cups}

\textbf{Filename:} stack\_cups.py

\textbf{Task:} Stack all cups on top of the specified color cup. The cup colors are sampled from the full set of 20 color instances. The scene always contains three cups.

\textbf{New/Modified:} No.

\textbf{Objects:} 3 tall cups.

\textbf{Variations per level:} Level 4: 0 (red) and 3 (green).

\textbf{Success Metric:}  All other cups are inside the specified cup.
\newline

\noindent\textbf{Put All Groceries In Cupboard}

\textbf{Filename:} put\_all\_groceries\_in\_cupboard.py

\textbf{Task:} Grab each one of the groceries and put them inside the cupboard until all of them are inside the cupboard.

\textbf{New/Modified:} Yes, modified to add new instructions and improve the groceries names.

\textbf{Objects:} 6 YCB objects (1 crackers box, 1 chocolate jello box, 1 strawberry jello box, 1 soup can, 1 spam can, 1 mustard bottle and 1 sugar box), and 1 cupboard (that hovers in the air like magic).

\textbf{Variations per level:} Level 4: 0.

\textbf{Success Metric:}  All the groceries are set inside the cupboard.

\end{document}